\documentclass[sigconf]{acmart}
\AtBeginDocument{%
  \providecommand\BibTeX{{%
    \normalfont B\kern-0.5em{\scshape i\kern-0.25em b}\kern-0.8em\TeX}}}

\setcopyright{acmlicensed}
\copyrightyear{2018}
\acmYear{2018}
\acmDOI{XXXXXXX.XXXXXXX}
\acmConference[Conference acronym 'XX]{Make sure to enter the correct
  conference title from your rights confirmation emai}{June 03--05,
  2018}{Woodstock, NY}
\acmISBN{978-1-4503-XXXX-X/18/06}

\usepackage{subcaption}
\usepackage{colortbl}
\definecolor{zhz_gray}{rgb}{0.8,0.8,0.8}

\newcommand{\nop}[1]{}

\begin{document}

\title{One Train for Two Tasks: An Encrypted Traffic Classification Framework Using Supervised Contrastive Learning}

\author{Haozhen Zhang}
\affiliation{%
  \institution{Shenzhen International Graduate School, Tsinghua University}
  \country{China}
  }
\email{zhang-hz21@mails.tsinghua.edu.cn}

\author{Xi Xiao}
\authornote{Corresponding author.}
\affiliation{%
  \institution{Shenzhen International Graduate School, Tsinghua University}
  \country{China}
  }
\email{xiaox@sz.tsinghua.edu.cn}

\author{Le Yu}
\affiliation{%
  \institution{Nanjing University of Posts and Telecommunications}
  \country{China}
  }
\email{yulele08@njupt.edu.cn}

\author{Qing Li}
\affiliation{%
  \institution{Peng Cheng Laboratory}
  \country{China}
  }
\email{liq@pcl.ac.cn}

\author{Zhen Ling}
\affiliation{%
  \institution{Southeast University}
  \country{China}
  }
\email{zhenling@seu.edu.cn}

\author{Ye Zhang}
\affiliation{%
  \institution{National University of Singapore}
  \country{Singapore}
  }
\email{e1124294@u.nus.edu}

\begin{abstract}

As network security receives widespread attention, encrypted traffic classification has become the current research focus. However, existing methods conduct traffic classification without sufficiently considering the common characteristics between data samples, leading to suboptimal performance. 
Moreover, they train the packet-level and flow-level classification tasks independently, which is redundant because the packet representations learned in the packet-level task can be exploited by the flow-level task. Therefore, in this paper, we propose an effective model named a \textbf{C}ontrastive \textbf{L}earning \textbf{E}nhanced \textbf{T}emporal \textbf{F}usion \textbf{E}ncoder (\textbf{CLE-TFE}). In particular, we utilize supervised contrastive learning to enhance the packet-level and flow-level representations and perform graph data augmentation on the byte-level traffic graph so that the fine-grained semantic-invariant characteristics between bytes can be captured through contrastive learning. 
We also propose cross-level multi-task learning, which simultaneously accomplishes the packet-level and flow-level classification tasks in the same model with one training. 
Further experiments show that CLE-TFE achieves the best overall performance on the two tasks, while its computational overhead (i.e., floating point operations, FLOPs) is only about 1/14 of the pre-trained model (e.g., ET-BERT). 
We release the code at \url{https://github.com/ViktorAxelsen/CLE-TFE}.

\end{abstract}



\maketitle

\section{Introduction}
\label{sec:intro}

As advancements in computer network technology continue and a large number of devices connect to the Internet, user privacy becomes increasingly vulnerable to malicious attacks. 
While encryption technologies like VPNs and Tor~\cite{VPNTor} offer protection to users~\cite{Encryption}, they can paradoxically serve as tools for attackers to conceal their identities. Traditional data packet inspection (DPI) methods have lost effectiveness against encrypted traffic~\cite{EncrySurvey}. Designing a universally effective method to classify an attacker's network activities—such as website browsing or application usage—from encrypted traffic remains a formidable challenge.

In the past few years, many methods have been proposed to enhance the capability of encrypted traffic classification techniques. 
Among them, statistic-based methods~\cite{AppScanner, KFP, FlowPrint, CUMUL, ETC-PS} generally rely on hand-crafted traffic statistical features and then leverage a traditional machine learning model for classification. 
However, they require heavy feature engineering and are susceptible to unreliable flows~\cite{TFE-GNN}. 
With the burgeoning of representation learning~\cite{CLSurvey}, some methods also use deep learning models to conduct traffic classification, such as pre-trained language models~\cite{ETBERT, PacRep}, neural networks~\cite{FSNet, TFE-GNN}, etc. 
For several samples with the same data label, there usually exist some common characteristics between them. 
Nevertheless, these methods directly learn representations of statistical features or raw bytes, without considering the potential commonalities between features of different samples, thus cannot sufficiently uncover the semantic-invariant information contained in the data. 
\textbf{Therefore, how to leverage common features between data to assist the model in learning more robust representations is a difficult problem.} 
Additionally, current methods cannot simultaneously train the packet-level and flow-level traffic classification tasks on the same model. So, they require at least two independent training for the flow-level and packet-level tasks, respectively. 
This is inconvenient and redundant because an informative packet representation has been learned while training on the packet-level task, and there is no need to learn it again from scratch on the flow-level task. 
\textbf{Consequently, how to utilize the potential relation between the two tasks of different levels in model training to improve model performance is also a vital challenge.}

To address the above challenges, in this paper, we propose a novel and simple yet effective model named a \textbf{C}ontrastive \textbf{L}earning \textbf{E}nhanced \textbf{T}emporal \textbf{F}usion \textbf{E}ncoder (\textbf{CLE-TFE}) for encrypted traffic classification. 
We build CLE-TFE based on TFE-GNN~\cite{TFE-GNN}, and CLE-TFE consists of two modules: the contrastive learning module and the cross-level multi-task learning module. 
The contrastive learning module conducts contrastive learning at both the packet and flow levels. 
The packet-level contrastive learning skillfully performs graph data augmentation on the byte-level traffic graph of TFE-GNN to better capture fine-grained semantic-invariant representations between bytes, leading to a robust packet-level representation. 
On top of this, the flow-level contrastive learning further augments packets in the flow to strengthen the flow-level representation. 
In particular, we use supervised contrastive learning~\cite{SCL} instead of unsupervised contrastive learning~\cite{CPC} to learn common features between samples with the same data label and further improve model performance. 
In addition, in the cross-level multi-task learning module, we use the same model to finish both the packet-level and the flow-level traffic classification tasks in one training and reveal the cross-level relation between the packet-level and the flow-level tasks: the packet-level task is helpful for the flow-level task. 
In the experiments, we adopt the ISCX VPN-nonVPN~\cite{ISCX} and the ISCX Tor-nonTor~\cite{ISCX-Tor} datasets with 20+ baselines to conduct a comprehensive evaluation of CLE-TFE on both the packet-level and the flow-level traffic classification tasks. 
Elaborately designed experiments and results demonstrate that CLE-TFE achieves the best overall performance on the two tasks of different levels. 
Note that compared with TFE-GNN, CLE-TFE adds almost no additional parameters, and the computational costs are significantly reduced by nearly half while achieving better model performance (2.4\%↑ and 5.7\%↑ w.r.t. f1-score on the ISCX-VPN and ISCX-nonTor datasets, respectively).

To summarize, our main contributions include: 
\begin{itemize}
\item To fully exploit common characteristics between data samples, we propose a simple yet effective model (i.e., CLE-TFE) that utilizes supervised contrastive learning. 
We perform graph data augmentation on the byte-level traffic graph to uncover fine-grained semantic-invariant representations between bytes through contrastive learning. 
\item To our best knowledge, we are the \textit{first} to accomplish both the packet-level and flow-level traffic classification tasks in the same model with one training through cross-level multi-task learning with almost no additional parameters. 
We also find that the packet-level task benefits the flow-level task. 
\item We conduct experiments on both the packet-level and flow-level classification tasks using the ISCX dataset~\cite{ISCX, ISCX-Tor}. The experimental results show that CLE-TFE achieves the overall best performance. 
\end{itemize}

\section{Preliminaries}
\label{sec:preliminaries}

\textbf{Encrypted Traffic Classification.}
Encrypted traffic classification aims to identify the traffic source based on packet information captured by professional software or programs~\cite{ETCSurvey}. 
In this paper, we only utilize raw bytes for classification and mainly focus on the packet-level and flow-level classification tasks. 
Given a training dataset with $N_f$ traffic flows (defined by its five-tuple: the source and destination IP address, the source and destination ports, and protocol), $N_p$ packets, and $C$ categories in total, the packet-level classification task aims to classify each unseen test packet sample into a predicted category $\boldsymbol{y'}^p$ ($\boldsymbol{y}^p$ is the packet ground truth) with a well-trained model $S_p$ on $N_p$ packet training samples, where $\boldsymbol{y'}^p = 0, 1, \cdots, C-1$. 
Similarly, the flow-level classification task aims to classify each unseen test flow sample into a predicted category $\boldsymbol{y'}^f$ ($\boldsymbol{y}^f$ is the flow ground truth) with another well-trained model $S_f$ on $N_f$ flow training samples, where $\boldsymbol{y'}^f = 0, 1, \cdots, C-1$. 
Note that the packet ground truth $\boldsymbol{y}^p$ is consistent with the ground truth $\boldsymbol{y}^f$ of the flow it belongs to. 
In this paper, we perform only one training on the same model to simultaneously accomplish both packet-level and flow-level classification tasks, making $S_p$ and $S_f$ completely consistent in architecture and parameters (i.e., $S_p \equiv S_f$).

\textbf{Temporal Fusion Encoder.}
The temporal fusion encoder (TFE-GNN)~\cite{TFE-GNN} is a flow-level traffic classification model. 
It utilizes point-wise mutual information~\cite{PMI} to construct the (undirected) byte-level traffic graph $\mathcal{G} = \{\mathcal{V}, \mathcal{E}\}$ for each packet, where $\mathcal{V}$ is the node set denoting byte values and $\mathcal{E}$ is the edge set denoting the semantic correlation between bytes (for simplicity, we refer to the header and payload byte-level traffic graph of TFE-GNN collectively as the traffic graph). 
TFE-GNN designs a traffic graph encoder to encode the traffic graph $\mathcal{G}$ into the packet-level representation $\mathbf{p}$, which can be described as:
\begin{equation}
\mathbf{p} = \operatorname{TFE-GNN}(\mathcal{G})
\end{equation}
TFE-GNN further leverages long-short term memory (LSTM)~\cite{LSTM} to obtain the flow-level representation $\mathbf{f}$: 
\begin{equation}
\mathbf{f} = \operatorname{LSTM}(\mathbf{p}_1, \mathbf{p}_2, \cdots, \mathbf{p}_n)
\end{equation}
where $n$ is the flow length, $\mathbf{p}_n$ is the $n$-th packet-level representation in the flow. 
We build our model based on TFE-GNN in this paper.

\textbf{Contrastive Learning.}
Contrastive learning is a widely used representation learning method in computer vision that aims to learn semantic-invariant representations by contrasting positive and negative sample pairs~\cite{tripletloss, CPC, MoCo, SimCLR}. 
Generally, most contrastive learning methods employ various data augmentation on the original data (e.g., color jitter, random flop) to construct two different augmented views, which are further encoded by a shared encoder. 
Moreover, some methods apply contrastive learning to graph representation learning, which leverages graph data augmentation (GDA) to construct the augmented view~\cite{DGI, GraphCL, GCC, GCA}. 
Given a minibatch of $N$ samples during training, we can obtain two augmented views containing $2N$ augmented samples through data augmentation. 
Using the shared encoder, we further obtain the embedding vector $\mathbf{z}$ for each augmented sample. 
In particular, an unsupervised contrastive loss~\cite{CPC, GraphCL} is utilized to maximize agreement between the two augmented views, which can be formulated as: 
\begin{equation}
\label{eq:ucl}
\mathcal{L}_{\text{CL}}=-\sum_{i \in I} \log \frac{\exp \left(\mathbf{z}_i \cdot \mathbf{z}_{j(i)} / \tau\right)}{\sum_{k \in K(i)} \exp \left(\mathbf{z}_i \cdot \mathbf{z}_k / \tau\right)}
\end{equation}
where $i \in I \equiv\{1 \ldots 2 N\}$ is the index of an arbitrary augmented sample, $j(i)$ (the \textit{positive}) is the index of another augmented sample from the same original sample as $i$, and $K(i) \equiv I \backslash\{i\}$ (Note that $I \backslash\{i, j(i)\}$ is the \textit{negatives}). 
The symbol $\cdot$ represents the inner product operation and $\tau \in \mathcal{R}^{+}$ denoted the temperature parameter. 
In this way, similar samples (i.e., the positive sample pairs) are closer in the embedding space, and dissimilar samples (i.e., the negative sample pairs) are further apart, resulting in a better representation.

\begin{figure*}[htb]
	\centering
	\includegraphics[width=0.95\linewidth]{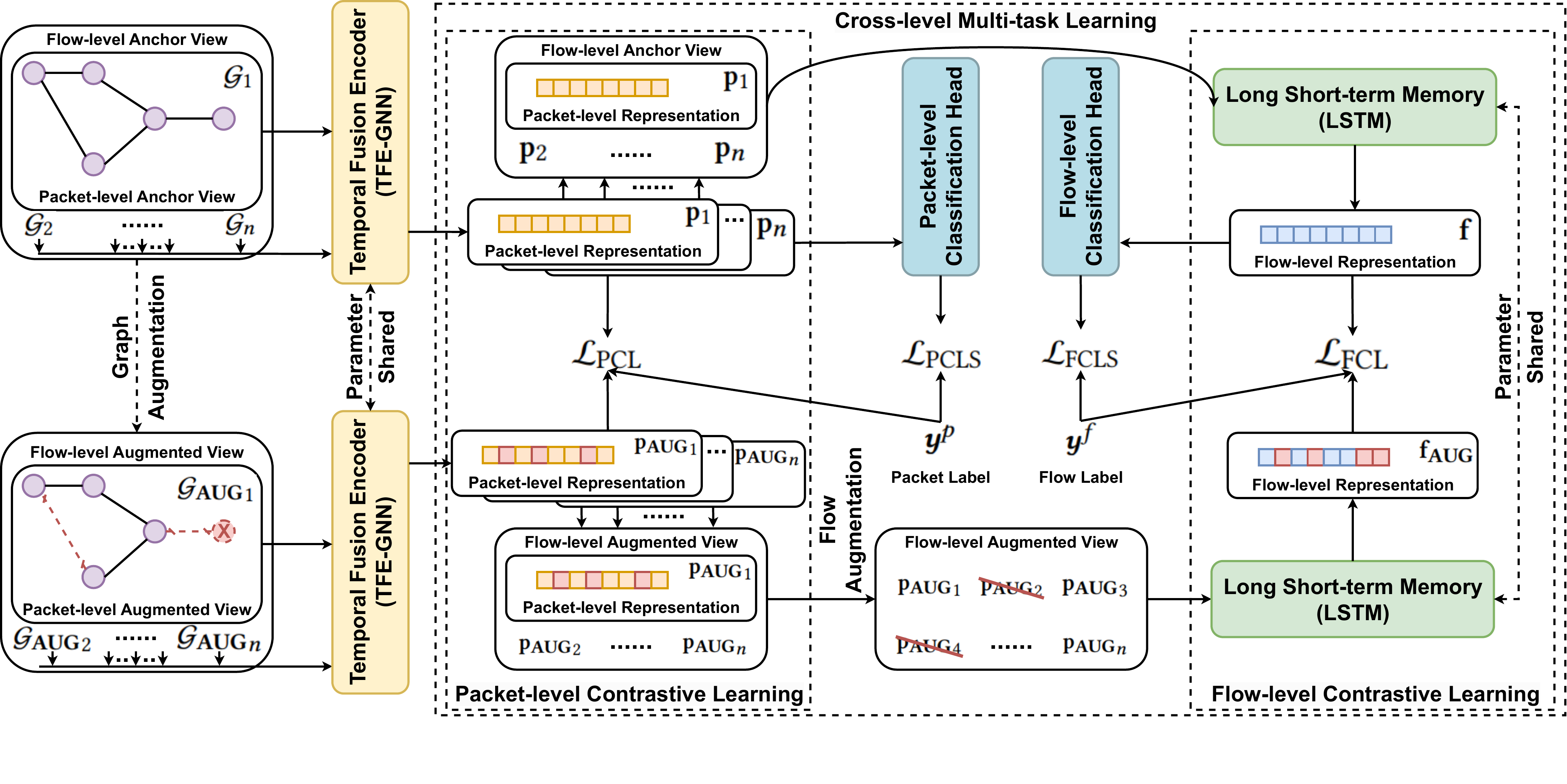}
	\vspace{-5ex}
        \caption{CLE-TFE Model Architecture}
        \vspace{-1ex}
	\label{model_figure}
\end{figure*}

\section{Methodology}
\label{sec:method}

\subsection{Framework Overview}
\label{sec:method_motivation}

As shown in Figure \ref{model_figure}, our model consists of two main modules: the \textit{contrastive learning module} (Section~\ref{sec:method_cl}) and the \textit{cross-level multi-task learning module} (Section~\ref{sec:method_mtl}). 
The contrastive learning module contains the packet-level and flow-level contrastive learning modules. The packet-level contrastive learning module applies graph data augmentation to the traffic graph to construct a packet-level augmented view for packet-level contrastive learning. The flow-level contrastive learning module further constructs a flow-level augmented view by randomly removing packets in the flow, which is used for flow-level contrastive learning. 
The cross-level multi-task learning module leverages the relation between the packet-level and flow-level tasks to jointly train the two levels' contrastive learning and classification tasks for better representations.

\subsection{Contrastive Learning at Dual Levels}
\label{sec:method_cl}

Inspired by the powerful representation learning capabilities of contrastive learning (Section~\ref{sec:preliminaries}), we propose to utilize it to enhance the packet-level and flow-level representations in TFE-GNN~\cite{TFE-GNN}, which will be detailed in the following two sections.

\subsubsection{Packet-level Contrastive Learning}
\label{sec:method_pcl}

Since TFE-GNN uses raw bytes to construct the traffic graph, we need to construct the packet-level augmented view by perturbing the byte sequence, which is not a convenient and direct way. 
Instead, since the nodes and edges on the traffic graph represent byte values and the semantic correlation between bytes, respectively, directly perturbing nodes or edges on the traffic graph has a finer granularity and can better uncover semantic-invariant representations between bytes through contrastive learning, so we choose to use graph data augmentation to perturb and construct the packet-level augmented view on the traffic graph~\cite{GraphCL}, which includes node dropping and edge dropping. 

\textbf{Node Dropping}. Given the traffic graph $\mathcal{G}$, we randomly drop nodes along with their connecting edges with a certain probability, which is equivalent to deleting the corresponding bytes in the original byte sequence. Node dropping can be formulated as: 
\begin{equation}
\mathcal{G'} = \left\{\left\{v_i \odot \rho_i \mid v_i \in \mathcal{V}\right\}, \left\{e_{i j} \odot \rho_i \mid e_{i j} \in \mathcal{E}\right\}\right\} = \{\mathcal{V'}, \mathcal{E'}\}
\end{equation}
where $\rho_i \in\{0,1\}$ is drawn from a Bernoulli distribution $\rho_i \sim \mathcal{B}(P_{\text{ND}})$, which denotes whether to drop node $v_i$ with its connecting edges $e_{i j}$. $P_{\text{ND}} \in [0, 1]$ is the node dropping ratio, which is a hyper-parameter controlling the expected amount of dropped nodes. 

\textbf{Edge Dropping}. We randomly drop edges on the node-dropping graph $\mathcal{G'}$, which is equivalent to removing the semantic correlation between two arbitrary bytes in the original byte sequence. Edge dropping can be formulated as: 
\begin{equation}
\mathcal{G}_{\text{AUG}} = \left\{\mathcal{V'},\left\{e_{i j} \odot \rho_{i j} \mid e_{i j} \in \mathcal{E'}\right\}\right\}
\end{equation}
where $\rho_{i j} \in\{0,1\}$ obeys $\rho_{i j} \sim \mathcal{B}(P_{\text{ED}})$, denoting whether to drop edge $e_{i j}$. $P_{\text{ED}} \in [0, 1]$ is the edge dropping ratio, which is a hyper-parameter controlling the expected amount of dropped edges.

After the augmentations, we obtain the packet-level augmented view $\mathcal{G}_{AUG}$ and its embedding vector: 
\begin{equation}
\mathbf{p}_{\text{AUG}} = \operatorname{TFE-GNN}(\mathcal{G}_{AUG})
\end{equation}
Notably, we treat the original traffic graph $\mathcal{G}$ as a packet-level anchor view without augmentation for training stability. 
In Eq.~\ref{eq:ucl}, the unsupervised contrastive loss only treats two different views of the same sample as positive sample pairs, which is insufficient. 
Instead, we propose to utilize supervised contrastive loss~\cite{SCL} to take advantage of the data labels during training. 
Given a sample in a minibatch, the supervised contrastive loss treats all other samples with the same data label as positive samples. This significantly increases the number of positive sample pairs and can bring similar samples closer in the embedding space, yielding a more robust representation. 
By incorporating supervision in Eq.~\ref{eq:ucl}, the packet-level supervised contrastive loss can be formulated as: 
\begin{equation}
\mathcal{L}_{\text{PCL}}=\sum_{i \in I} \frac{-1}{|M(i)|} \sum_{m \in M(i)} \log \frac{\exp \left(\mathbf{p'}_i \cdot \mathbf{p'}_m / \tau\right)}{\sum_{k \in K(i)} \exp \left(\mathbf{p'}_i \cdot \mathbf{p'}_k / \tau\right)}
\end{equation}
where $\mathbf{p'}$ refers to the embedding vector collection of $\mathbf{p}_{\text{AUG}}$ and $\mathbf{p}$. 
$M(i) \equiv\left\{m \in K(i): \boldsymbol{y}^p_m=\boldsymbol{y}^p_i\right\}$ is the indices of all positive samples of $i$. 
The negative sample selection remains the same as Eq.~\ref{eq:ucl} because the performance generally becomes better with increasing negatives, as many works suggest~\cite{MoCo, SimCLR, DEIR, CMC}.

\subsubsection{Flow-level Contrastive Learning}
\label{sec:method_fcl}

We aim to enhance the model performance further by conducting the flow-level contrastive learning task. 
Based on the packet-level augmented view $\mathcal{G}_{AUG}$ and its embedding vector $\mathbf{p}_{\text{AUG}}$, we augment the traffic flow by randomly dropping packets from it, which is equivalent to simulating network packet loss and can be formulated as: 
\begin{equation}
\mathbf{f_{\text{AUG}}} = \operatorname{LSTM}({\mathbf{p}_{\text{AUG}}}_1 \odot \rho_1, {\mathbf{p}_{\text{AUG}}}_2 \odot \rho_2, \cdots, {\mathbf{p}_{\text{AUG}}}_n \odot \rho_n)
\end{equation}
where we feed the flow-level augmented view $({\mathbf{p}_{\text{AUG}}}_1 \odot \rho_1, {\mathbf{p}_{\text{AUG}}}_2 \odot \rho_2, \cdots, {\mathbf{p}_{\text{AUG}}}_n \odot \rho_n)$ into LSTM and obtain its embedding vector $\mathbf{f_{\text{AUG}}}$. 
$\rho_i \in\{0,1\}$ obeys $\rho_i \sim \mathcal{B}(P_{\text{PD}})$, which denotes whether to drop packet ${\mathbf{p}_{\text{AUG}}}_i$. $P_{\text{PD}} \in [0, 1]$ is the packet dropping ratio, which is a hyper-parameter controlling the expected amount of dropped packets in a flow. 
The flow-level representation $\mathbf{f}$ also acts as the embedding vector of the flow-level anchor view $(\mathbf{p}_1, \mathbf{p}_2, \cdots, \mathbf{p}_n)$ for stable training. 
Similarly, we also adopt the supervised contrastive loss~\cite{SCL} in the flow-level contrastive learning, which can be formulated as: 
\begin{equation}
\mathcal{L}_{\text{FCL}}=\sum_{i \in I} \frac{-1}{|N(i)|} \sum_{n \in N(i)} \log \frac{\exp \left(\mathbf{f'}_i \cdot \mathbf{f'}_n / \tau\right)}{\sum_{k \in K(i)} \exp \left(\mathbf{f'}_i \cdot \mathbf{f'}_k / \tau\right)}
\end{equation}
where $\mathbf{f'}$ refers to the embedding vector collection of $\mathbf{f}_{\text{AUG}}$ and $\mathbf{f}$. 
$N(i) \equiv\left\{n \in K(i): \boldsymbol{y}^f_n=\boldsymbol{y}^f_i\right\}$ is the indices of all positive samples of $i$. 
Such learning paradigms can also help the model capture the common characteristics of the traffic flow through augmentation, leading to a robust flow-level representation.

\subsection{Cross-level Multi-task Learning}
\label{sec:method_mtl}
\textbf{The Packet-level Classification Task Extension.}
Because TFE-GNN~\cite{TFE-GNN} can encode each packet into a high-dimensional vector independently, we can easily extend it to the packet-level classification task based on the flow-level classification task. 
As shown in Figure \ref{model_figure}, we feed the high-dimensional packet vectors $\mathbf{p}$ into an additional packet-level classification head, which is a multi-layer perception (MLP), to conduct the packet-level classification task.

\noindent \textbf{Cross-level Relation.}
Since a flow is composed of several packets, a better packet representation should facilitate learning the flow representation.
Using the original design of TFE-GNN~\cite{TFE-GNN}, the traffic graph encoder, which is responsible for encoding packet vectors $\mathbf{p}$, can only be optimized by receiving cross-level supervision signals from the flow-level classification and contrastive tasks.
However, the cross-level supervision signals from flow-level tasks cannot sufficiently optimize the packet vectors $\mathbf{p}$, resulting in suboptimal performance. 
In other words, packet vectors lack supervision signals that can optimize them directly. 
Intuitively, the packet-level classification and contrastive tasks can naturally play this role, which gives the packet vectors $\mathbf{p}$ an additional constraint in their embedding space so that the packet vectors $\mathbf{p}$ can be optimized more sufficiently, leading to a better flow representation $\mathbf{f}$. 
Therefore, leveraging this cross-level relation, we can simultaneously perform multi-task learning on both the flow-level and packet-level tasks, which unifies the two classical traffic classification tasks into one model with one training. 
The cross-level multi-task learning module also improves the overall model performance, which is mutually beneficial.

\subsection{Training Objective}
\label{sec:method_objective}

We first give the formal formula description of the flow-level and the packet-level classification task. 
To conduct the flow-level classification task, we use the flow-level classification head to perform a non-linear transformation on the flow-level representation $f$: 
\begin{equation}
\mathbf{\hat{f}} = \mathbf{w}_{f2}^T\operatorname{PReLU}(\mathbf{w}_{f1}^T \mathbf{f} + \mathbf{b}_{f1}) + \mathbf{b}_{f2}
\end{equation}
where PReLU~\cite{PReLU} is the activation function. $\mathbf{w}_{f1}, \mathbf{w}_{f2}$ and $\mathbf{b}_{f1}, \mathbf{b}_{f2}$ are the weights and biases of the flow-level classification head. 
The flow-level classification task can be further described as: 
\begin{equation}
\mathcal{L}_{\text{FCLS}} = \operatorname{CE}(\mathbf{\hat{f}}, \boldsymbol{y}^f)
\end{equation}
where $\operatorname{CE}(\cdot)$ is the cross entropy loss function. 
Similarly, the packet-level classification task can be written as: 
\begin{equation}
\mathbf{\hat{p}} = \mathbf{w}_{p2}^T\operatorname{PReLU}(\mathbf{w}_{p1}^T \mathbf{p} + \mathbf{b}_{p1}) + \mathbf{b}_{p2}
\end{equation}
\begin{equation}
\mathcal{L}_{\text{PCLS}} = \operatorname{CE}(\mathbf{\hat{p}}, \boldsymbol{y}^p)
\end{equation}
where $\boldsymbol{y}^p$ is consistent with the flow label it belongs to. $\mathbf{w}_{p1}, \mathbf{w}_{p2}$ and $\mathbf{b}_{p1}, \mathbf{b}_{p2}$ are the weights and biases of the packet-level classification head. 
In summary, we propose the overall end-to-end training objective of CLE-TFE as follows: 
\begin{equation}
\mathcal{L} = \mathcal{L}_{\text{PCLS}} + \mathcal{L}_{\text{FCLS}} + \alpha \mathcal{L}_{\text{PCL}} + \beta \mathcal{L}_{\text{FCL}}
\end{equation}
where $\alpha, \beta \in [0, 1]$ are the coefficients that control the contribution of the packet-level and flow-level contrastive tasks, respectively.

In this way, we can complete the flow-level and packet-level tasks in the same model with one training. 
The trained model can be directly leveraged to evaluate the flow-level and packet-level classification tasks without two independent training, which is convenient and reduces computing resource consumption. 
We will further analyze the computational costs of CLE-TFE in Section~\ref{sec:exp_compute}.

\section{Experiments}
\label{sec:exp}

\subsection{Experimental Settings}
\label{sec:exp_setttings}

\subsubsection{Dataset}

To thoroughly evaluate the superiority of CLE-TFE on the packet-level and flow-level traffic classification tasks, we select the ISCX VPN-nonVPN~\cite{ISCX} and ISCX Tor-nonTor~\cite{ISCX-Tor} datasets. 
The ISCX VPN-nonVPN dataset is a collection of two datasets: the ISCX-VPN and ISCX-nonVPN datasets. The ISCX-VPN dataset contains traffic collected over virtual private networks (VPNs), which are commonly used for accessing blocked websites or services. Due to obfuscation technology, this kind of traffic can be challenging to detect. In contrast, the ISCX-nonVPN dataset contains regular traffic not collected over VPNs. 
The ISCX Tor-nonTor dataset consists of the ISCX-Tor and ISCX-nonTor datasets. The ISCX-Tor dataset involves traffic collected over the onion router (Tor), making its traffic difficult to trace, whereas the ISCX-nonTor is a regular dataset not collected over Tor. 

In the experiment, we divide the traffic data in the ISCX VPN-nonVPN and ISCX Tor-nonTor datasets into six and eight categories according to the type of traffic in the datasets following~\cite{TFE-GNN}. We conduct all experiments independently on these four datasets (i.e., ISCX-VPN, ISCX-NonVPN, ISCX-Tor, and ISCX-NonTor). 
The dataset statistics and category details are given in Appendix \ref{sec:appendix_dataset}, and we describe the threat model and assumptions in Appendix \ref{sec:appendix_threat}.

\subsubsection{Pre-processing}

For each dataset, we use SplitCap to obtain bidirectional flows from each pcap file. 
Due to the limited number of flows in the ISCX-Tor dataset, we enrich traffic flows by dividing each flow into 60-second non-overlapping blocks in our experiments~\cite{FlowPic}. 
Following~\cite{TFE-GNN}, we filter out traffic flows without payload or that surpass a length of 10000. Such flows are usually used to establish connections between two communicating entities or result from temporary network failures, which have little useful information for classification. As for each packet in a traffic flow,  we first remove bad packets and retransmission packets. Then, we remove the Ethernet header, IP addresses, and port numbers to protect sensitive information while eliminating the potential interference it brings. After the above processing, we only keep the first 15 packets of a traffic flow at most to ensure relatively smaller computation costs, which is enough to achieve good performance.

Since our method needs to perform both the flow-level and packet-level classification tasks in one training, we first adopt stratified sampling to partition the flow-level training and testing dataset into 9:1 according to the number of traffic flows for all datasets. All packets in the flow-level training and testing datasets are directly used as the packet-level training and testing datasets, respectively. The category of each packet is consistent with that of the traffic flow it belongs to. Note that independent packets can also be used to evaluate the packet-level task. Here, for convenience, the packets in a traffic flow are directly used for evaluation.

\subsubsection{Implementation Details and Baselines}

We use TFE-GNN~\cite{TFE-GNN} as the packet representation encoder. 
We set the max packet number within a flow (i.e., flow length) to 15. The edge dropping ratio $P_{\text{ED}}$ and node dropping ratio $P_{\text{ND}}$ are set to 0.05 and 0.1, respectively. The packet dropping ratio $P_{\text{PD}}$ is 0.6, and the temperature coefficient $\tau$ is 0.07. 
Other hyper-parameters depend on the specific dataset, and we detail them in Appendix \ref{sec:appendix_hyper}. 
We implement CLE-TFE and conduct all experiments with PyTorch and Deep Graph Library. The experimental results are reported as the mean over five runs on an NVIDIA RTX 3080 GPU.

As for evaluation metrics, we use Overall Accuracy (AC), Precision (PR), Recall (RC), and Macro F1-score (F1). We compare CLE-TFE with the flow-level and packet-level methods for a comprehensive comparison. The comparison baselines include \textbf{Flow-level Traffic Classification Methods} (i.e., AppScanner~\cite{AppScanner}, K-FP (K-Fingerprinting)~\cite{KFP}, FlowPrint~\cite{FlowPrint}, CUMUL~\cite{CUMUL}, GRAIN~\cite{GRAIN}, FAAR~\cite{FAAR}, ETC-PS~\cite{ETC-PS}, FS-Net~\cite{FSNet}, DF~\cite{DF}, EDC~\cite{EDC}, FFB~\cite{FFB}, MVML~\cite{MVML}, ET-BERT~\cite{ETBERT}, GraphDApp~\cite{GraphDApp}, ECD-GNN~\cite{ECD-GNN}, TFE-GNN~\cite{TFE-GNN}) and \textbf{Packet-level Traffic Classification Methods} (i.e., Securitas~\cite{Securitas}, 2D-CNN~\cite{2DCNN}, 3D-CNN~\cite{3DCNN}, DeepPacket (DP)~\cite{DeepPacket}, BLJAN~\cite{BLJAN}, EBSNN~\cite{EBSNN}).

For the rest of the experiment section, we will evaluate CLE-TFE from the following research questions: 

\noindent \textbf{RQ1}: How does CLE-TFE perform on the packet-level and flow-level tasks? (Section~\ref{sec:exp_results})

\noindent \textbf{RQ2}: How much does each module of CLE-TFE contribute to the model performance? (Section~\ref{sec:exp_ablation})

\noindent \textbf{RQ3}: How discriminative are the learned packet-level and flow-level representations in the embedding space? (Section~\ref{sec:exp_representation})

\noindent \textbf{RQ4}: How computationally expensive is CLE-TFE? (Section~\ref{sec:exp_compute})

\noindent \textbf{RQ5}: How sensitive is CLE-TFE to hyper-parameters? (Section~\ref{sec:exp_sensitivity})

\subsection{Comparison Experiments (RQ1)}
\label{sec:exp_results}

\begin{table*}[t]
  \caption{Experimental Results on ISCX VPN-nonVPN and ISCX Tor-nonTor Datasets w.r.t. Flow-level Classification Task. (Partial Results Are Taken From~\cite{TFE-GNN})}
  \label{tab:ISCXresults}
  \vspace{-1ex}
  \resizebox{\linewidth}{!}{
  \begin{tabular}{c|cccc|cccc|cccc|cccc}
    \toprule
    Dataset & \multicolumn{4}{c|}{ISCX-VPN} & \multicolumn{4}{c|}{ISCX-nonVPN} & \multicolumn{4}{c|}{ISCX-Tor} & \multicolumn{4}{c}{ISCX-nonTor} \\
    \midrule
    Model & AC & PR & RC & F1 & AC & PR & RC & F1 & AC & PR & RC & F1 & AC & PR & RC & F1 \\
    \midrule
    AppScanner~\cite{AppScanner}
    & 0.8889 & 0.8679 & 0.8815 & 0.8722
    & 0.7576 & 0.7594 & 7465 & 0.7486
    & 0.7543 & 0.6629 & 0.6042 & 0.6163
    & 0.9153 & 0.8435 & 0.8140 & 0.8273 \\
    K-FP~\cite{KFP}
    & 0.8713 & 0.8750 & 0.8748 & 0.8747
    & 0.7551 & 0.7478 & 0.7354 & 0.7387
    & 0.7771 & 0.7417 & 0.6209 & 0.6313
    & 0.8741 & 0.8653 & 0.7792 & 0.8167 \\
    FlowPrint~\cite{FlowPrint}
    & 0.8538 & 0.7451 & 0.7917 & 0.7566
    & 0.6944 & 0.7073 & 0.7310 & 0.7131
    & 0.2400 & 0.0300 & 0.1250 & 0.0484 
    & 0.5243 & 0.7590 & 0.6074 & 0.6153 \\
    CUMUL~\cite{CUMUL}
    & 0.7661 & 0.7531 & 0.7852 & 0.7644
    & 0.6187 & 0.5941 & 0.5971 & 0.5897
    & 0.6686 & 0.5349 & 0.4899 & 0.4997 
    & 0.8605 & 0.8143 & 0.7393 & 0.7627 \\
    GRAIN~\cite{GRAIN}
    & 0.8129 & 0.8077 & 0.8109 & 0.8027
    & 0.6667 & 0.6532 & 0.6664 & 0.6567
    & 0.6914 & 0.5253 & 0.5346 & 0.5234 
    & 0.7895 & 0.6714 & 0.6615 & 0.6613 \\
    FAAR~\cite{FAAR}
    & 0.8363 & 0.8224 & 0.8404 & 0.8291
    & 0.7374 & 0.7509 & 0.7121 & 0.7252
    & 0.6971 & 0.5915 & 0.4876 & 0.4814 
    & 0.9103 & 0.8253 & 0.7755 & 0.7959 \\
    ETC-PS~\cite{ETC-PS}
    & 0.8889 & 0.8803 & 0.8937 & 0.8851
    & 0.7273 & 0.7414 & 0.7133 & 0.7208
    & 0.7486 & 0.6811 & 0.5929 & 0.6033
    & 0.9365 & 0.8700 & 0.8311 & 0.8486 \\
    FS-Net~\cite{FSNet}
    & 0.9298 & 0.9263 & 0.9211 & 0.9234
    & 0.7626 & 0.7685 & 0.7534 & 0.7555
    & 0.8286 & 0.7487 & 0.7197 & 0.7242 
    & 0.9278 & 0.8368 & 0.8254 & 0.8285 \\
    DF~\cite{DF}
    & 0.8012 & 0.7799 & 0.8152 & 0.7921
    & 0.6742 & 0.6857 & 0.6717 & 0.6701
    & 0.6514 & 0.4803 & 0.4767 & 0.4719
    & 0.8568 & 0.8003 & 0.7415 & 0.7590 \\
    EDC~\cite{EDC}
    & 0.7836 & 0.7747 & 0.8108 & 0.7888
    & 0.6970 & 0.7153 & 0.7000 & 0.6978
    & 0.6400 & 0.4980 & 0.4528 & 0.4504 
    & 0.8692 & 0.7994 & 0.7411 & 0.7451 \\
    FFB~\cite{FFB}
    & 0.8304 & 0.8714 & 0.8149 & 0.8335
    & 0.7020 & 0.7274 & 0.6945 & 0.7050
    & 0.6343 & 0.4870 & 0.5203 & 0.4952 
    & 0.8954 & 0.7545 & 0.7430 & 0.7430 \\
    MVML~\cite{MVML}
    & 0.6491 & 0.7231 & 0.6198 & 0.6151
    & 0.5126 & 0.5751 & 0.4707 & 0.4806
    & 0.6343 & 0.3914 & 0.4104 & 0.3752 
    & 0.7235 & 0.5488 & 0.5512 & 0.5457 \\
    ET-BERT~\cite{ETBERT}
    & 0.9532 & 0.9436 & 0.9507 & 0.9463
    & 0.9167 & 0.9245 & 0.9229 & 0.9235
    & 0.9543 & 0.9242 & 0.9606 & 0.9397
    & 0.9029 & 0.8560 & 0.8217 & 0.8332 \\
    GraphDApp~\cite{GraphDApp}
    & 0.6491 & 0.5668 & 0.6103 & 0.5740
    & 0.4495 & 0.4230 & 0.3647 & 0.3614
    & 0.4286 & 0.2557 & 0.2509 & 0.2281
    & 0.6936 & 0.5447 & 0.5398 & 0.5352 \\
    ECD-GNN~\cite{ECD-GNN}
    & 0.1111 & 0.0185 & 0.1667 & 0.0333
    & 0.0606 & 0.0101 & 0.1667 & 0.0190
    & 0.0571 & 0.0071 & 0.1250 & 0.0135 
    & 0.9078 & 0.8015 & 0.8168 & 0.7977 \\
    TFE-GNN~\cite{TFE-GNN}
    & 0.9591 & 0.9526 & 0.9593 & 0.9536
    & 0.9040 & 0.9316 & 0.9190 & 0.9240
    & 0.9886 & 0.9792 & 0.9939 & 0.9855
    & 0.9390 & 0.8742 & 0.8335 & 0.8507 \\
    \midrule
    \textbf{CLE-TFE}
    & \textbf{0.9813} & \textbf{0.9771} & \textbf{0.9762} & \textbf{0.9761}
    & \textbf{0.9286} & \textbf{0.9396} & \textbf{0.9391} & \textbf{0.9389}
    & \textbf{1.0000} & \textbf{1.0000} & \textbf{1.0000} & \textbf{1.0000} 
    & \textbf{0.9554} & \textbf{0.9009} & \textbf{0.9019} & \textbf{0.8994} \\
    \bottomrule
  \end{tabular}
  }
\end{table*}
\begin{table*}[t]
  \caption{Experimental Results on ISCX VPN-nonVPN and ISCX Tor-nonTor Datasets w.r.t. Packet-level Classification Task. }
  \label{tab:ISCXresults_packet}
  \vspace{-1ex}
  \resizebox{\linewidth}{!}{
  \begin{tabular}{c|cccc|cccc|cccc|cccc}
    \toprule
    Dataset & \multicolumn{4}{c|}{ISCX-VPN} & \multicolumn{4}{c|}{ISCX-nonVPN} & \multicolumn{4}{c|}{ISCX-Tor} & \multicolumn{4}{c}{ISCX-nonTor} \\
    \midrule
    Model & AC & PR & RC & F1 & AC & PR & RC & F1 & AC & PR & RC & F1 & AC & PR & RC & F1 \\
    \midrule
    Securitas-C4.5~\cite{Securitas}
    & 0.8174 & 0.8128 & 0.8271 & 0.8194
    & 0.8833 & 0.8820 & 0.8842 & 0.8827
    & 0.8848 & 0.8669 & 0.8519 & 0.8577
    & 0.8475 & 0.8336 & 0.8501 & 0.8413 \\
    Securitas-SVM~\cite{Securitas}
    & 0.6447 & 0.6727 & 0.6488 & 0.6085
    & 0.7888 & 0.8096 & 0.7750 & 0.7817
    & 0.8244 & 0.7749 & 0.8320 & 0.7999
    & 0.7274 & 0.6970 & 0.7819 & 0.7327 \\
    Securitas-Bayes~\cite{Securitas}
    & 0.5796 & 0.6334 & 0.5703 & 0.5271
    & 0.7292 & 0.7974 & 0.6608 & 0.7061
    & 0.7681 & 0.7643 & 0.6500 & 0.6639
    & 0.6884 & 0.6868 & 0.6494 & 0.6384 \\
    2D-CNN~\cite{2DCNN}
    & 0.6595 & 0.7117 & 0.5601 & 0.5444
    & 0.4287 & 0.7166 & 0.4075 & 0.3653
    & 0.7641 & 0.5933 & 0.5814 & 0.5657
    & 0.4380 & 0.5710 & 0.3515 & 0.3521 \\
    3D-CNN~\cite{3DCNN}
    & 0.5480 & 0.6420 & 0.5428 & 0.4971
    & 0.6130 & 0.7153 & 0.6422 & 0.6223
    & 0.7096 & 0.8240 & 0.6498 & 0.6580
    & 0.6746 & 0.7438 & 0.4881 & 0.5062 \\
    DP-SAE~\cite{DeepPacket}
    & 0.7137 & 0.7187 & 0.6662 & 0.6777
    & 0.7241 & 0.7431 & 0.7626 & 0.7451
    & 0.7598 & 0.7791 & 0.6490 & 0.6567
    & 0.7156 & 0.6257 & 0.5850 & 0.5913 \\
    DP-CNN~\cite{DeepPacket}
    & 0.8508 & 0.8309 & 0.8307 & 0.8303
    & 0.7467 & 0.7597 & 0.7774 & 0.7669
    & 0.9388 & 0.9600 & 0.9347 & 0.9452
    & 0.8021 & 0.7360 & 0.6876 & 0.6971 \\
    BLJAN~\cite{BLJAN}
    & 0.6379 & 0.6423 & 0.7723 & 0.5957
    & 0.6914 & 0.6741 & 0.7727 & 0.6960
    & 0.9913 & 0.9953 & 0.9933 & 0.9942
    & 0.5856 & 0.5992 & 0.6599 & 0.5674 \\
    EBSNN-GRU~\cite{EBSNN}
    & \textbf{0.9588} & \textbf{0.9604} & 0.9569 & \textbf{0.9584}
    & 0.8266 & 0.8618 & 0.8537 & 0.8440
    & 0.9925 & 0.9880 & 0.9958 & 0.9918
    & 0.9120 & 0.8622 & 0.8398 & 0.8501 \\
    EBSNN-LSTM~\cite{EBSNN}
    & 0.9583 & 0.9592 & \textbf{0.9587} & \textbf{0.9584}
    & 0.8798 & 0.8991 & 0.9020 & 0.9000
    & 0.9992 & 0.9995 & 0.9988 & 0.9992
    & 0.9120 & 0.8646 & 0.8453 & 0.8532 \\
    \midrule
    \textbf{CLE-TFE}
    & 0.9480 & 0.9445 & 0.9425 & 0.9433
    & \textbf{0.8853} & \textbf{0.9039} & \textbf{0.9033} & \textbf{0.9027}
    & \textbf{0.9996} & \textbf{0.9997} & \textbf{0.9998} & \textbf{0.9997} 
    & \textbf{0.9240} & \textbf{0.8795} & \textbf{0.8544} & \textbf{0.8649} \\
    \bottomrule
  \end{tabular}
  }
\end{table*}

The comparison results of the flow-level task and the packet-level task on the ISCX datasets are shown in Table \ref{tab:ISCXresults} and \ref{tab:ISCXresults_packet}, respectively. 

\textbf{Flow-level Classification Results}. 
According to Table \ref{tab:ISCXresults}, CLE-TFE achieves the best results w.r.t. all the metrics on the ISCX datasets. 
Compared with traditional statistical feature methods, our approach surpasses them by a large margin due to the powerful expressive ability of graph neural networks. 
As for deep learning methods, our method still has obvious advantages. 
Among these methods, the performance of ET-BERT~\cite{ETBERT} is the closest to that of CLE-TFE, which benefits from its large number of parameters and the pretraining on large-scale datasets. 
But at the same time, its computational overhead is relatively large. 
Especially, CLE-TFE improves the performance of TFE-GNN~\cite{TFE-GNN} by a significant margin, which verifies the effectiveness of our elaborately designed contrastive learning scheme.

\textbf{Packet-level Classification Results}. 
From Table \ref{tab:ISCXresults_packet}, CLE-TFE reaches the best results on the ISCX-nonVPN, ISCX-Tor, and ISCX-nonTor datasets. 
Securitas~\cite{Securitas}, which is a traditional packet classification method, performs much worse than CLE-TFE on the encrypted traffic such as the ISCX-VPN and ISCX-Tor datasets, but only slightly worse than CLE-TFE on the ISCX-NonVPN and ISCX-NonTor datasets, which illustrates the superior representation capability of CLE-TFE on encrypted traffic. 
Especially on the ISCX-VPN dataset, CLE-TFE is slightly weaker than EBSNN~\cite{EBSNN} on the packet-level classification task. 
It's mainly because EBSNN~\cite{EBSNN} only conducts the packet-level classification task independently, reducing the potential negative impact brought by the flow-level classification task. 
However, it is more difficult for CLE-TFE to optimize the flow-level and packet-level tasks simultaneously in one training, so the results of the two tasks will have some trade-offs. 

From the overall perspective of the flow-level and packet-level tasks, CLE-TFE reaches the best comprehensive performance.

\subsection{Ablation Study (RQ2)}
\label{sec:exp_ablation}

\begin{table*}[t]
  \footnotesize
  \caption{Ablation Study of CLE-TFE on the ISCX-VPN Dataset}
  \label{tab:Ablation_study}
  \vspace{-1ex}
  \begin{tabular}{c|cccc|cccc|cccc}
    \toprule
    Tasks & \multicolumn{4}{c|}{Flow-level Classification Task} & \multicolumn{4}{c|}{Packet-level Classification Task} & \multicolumn{4}{c}{Average on the Two Tasks} \\
    \midrule
    Variants & AC & PR & RC & F1 & AC & PR & RC & F1 & AC & PR & RC & F1 \\
    \midrule
    w/o Flow-level Classification Loss
    & - & - & - & - 
    & 0.9539 & 0.9533 & 0.9492 & 0.9507
    & 0.4770 & 0.4767 & 0.4746 & 0.4754 \\
    w/o Flow-level Contrastive Loss
    & 0.9743 & 0.9744 & 0.9684 & 0.9710 
    & 0.9426 & 0.9386 & 0.9349 & 0.9364
    & 0.9585 & 0.9565 & 0.9517 & 0.9537  \\
    w/o Flow-level Classification \& Contrastive Loss
    & - & - & - & - 
    & \textbf{0.9548} & \textbf{0.9559} & \textbf{0.9498} & \textbf{0.9522}
    & 0.4774 & 0.4780 & 0.4749 & 0.4761 \\
    \midrule
    w/o Packet-level Classification Loss
    & 0.9696 & 0.9642 & 0.9624 & 0.9625 
    & - & - & - & - 
    & 0.4848 & 0.4821 & 0.4812 & 0.4763 \\
    w/o Packet-level Contrastive Loss
    & 0.9766 & 0.9738 & 0.9740 & 0.9735 
    & 0.9325 & 0.9312 & 0.9242 & 0.9271 
    & 0.9546 & 0.9525 & 0.9491 & 0.9503 \\
    w/o Packet-level Classification \& Contrastive Loss
    & 0.9474 & 0.9366 & 0.9373 & 0.9354 
    & - & - & - & - 
    & 0.4737 & 0.4683 & 0.4687 & 0.4677 \\
    \midrule
    w/o Packet-level Header Graph Aug
    & 0.9719 & 0.9692 & 0.9651 & 0.9669 
    & 0.9387 & 0.9343 & 0.9315 & 0.9328 
    & 0.9553 & 0.9518 & 0.9483 & 0.9499 \\
    w/o Packet-level Payload Graph Aug
    & 0.9790 & \textbf{0.9774} & 0.9751 & 0.9760 
    & 0.9422 & 0.9389 & 0.9358 & 0.9369 
    & 0.9606 & 0.9582 & 0.9555 & 0.9565 \\
    w/o Packet-level Header \& Payload Graph Aug
    & 0.9766 & 0.9736 & \textbf{0.9764} & 0.9748 
    & 0.9434 & 0.9382 & 0.9394 & 0.9384 
    & 0.9600 & 0.9559 & 0.9579 & 0.9566 \\
    \midrule
    w/ Unsupervised Contrastive Loss (No Label Used)
    & 0.9228 & 0.9208 & 0.9183 & 0.9189 
    & 0.8790 & 0.8725 & 0.8592 & 0.8651 
    & 0.9009 & 0.8967 & 0.8888 & 0.8920 \\
    \midrule
    \textbf{CLE-TFE}
    & \textbf{0.9813} & 0.9771 & 0.9762 & \textbf{0.9761} 
    & 0.9480 & 0.9445 & 0.9425 & 0.9433 
    & \textbf{0.9647} & \textbf{0.9608} & \textbf{0.9594} & \textbf{0.9597} \\
    \bottomrule
  \end{tabular}
\end{table*}

For a clear understanding of CLE-TFE architecture design, we conduct a comprehensive ablation study on the ISCX-VPN dataset, and the results are shown in Table \ref{tab:Ablation_study}. 
From Table \ref{tab:Ablation_study}, we can conclude that the packet-level and flow-level contrastive loss can improve the classification performance of the corresponding level task. 
However, jointly training the two-level tasks does not benefit them simultaneously. 
The packet-level classification \& contrastive tasks have a positive effect on the flow-level classification task, but conversely, the flow-level classification \& contrastive tasks have a slightly negative effect on the packet-level classification task. 
The underlying reason is that the packet-level classification \& contrastive tasks enable the model to learn a more discriminative packet-level representation, whereby the flow-level representation will also benefit from this and become more robust. 
However, the flow-level classification \& contrastive tasks optimize the flow-level representation, which cannot directly have a positive impact on the packet-level representation and may even slightly do harm to the learning of the packet-level representation. 
We will further analyze the learned representation of both levels in Section \ref{sec:exp_representation}. 

As for graph augmentation, both header and payload graph perturbations are conducive to improving the model performance, but the effect of header graph perturbation is slightly greater. 
This may be because the header contains crucial data packet attributes, while the payload only describes the content of that. 
The perturbation of attributes has a greater and more aggressive impact on the model performance than the perturbation of the content, so it can force the model to learn a more robust representation during the optimization process to resist the influence of such perturbation. 
Besides, we also give the results of using unsupervised contrastive learning loss (i.e., no label used), which is much worse than ours. 
It indicates that choosing the appropriate positive and negative sample pairs in contrastive learning can greatly improve the model's performance. 

In short, CLE-TFE achieves the best results compared to various variants w.r.t. the average performance of the packet-level and flow-level tasks. More ablation study results are in Appendix \ref{sec:appendix_ablation}.

\subsection{Analysis on the Learned Flow-level and Packet-level Representations (RQ3)}
\label{sec:exp_representation}

\begin{figure}[t]
        \centering
	\begin{subfigure}{0.49\linewidth}
		\centering
		\includegraphics[width=1.0\linewidth]{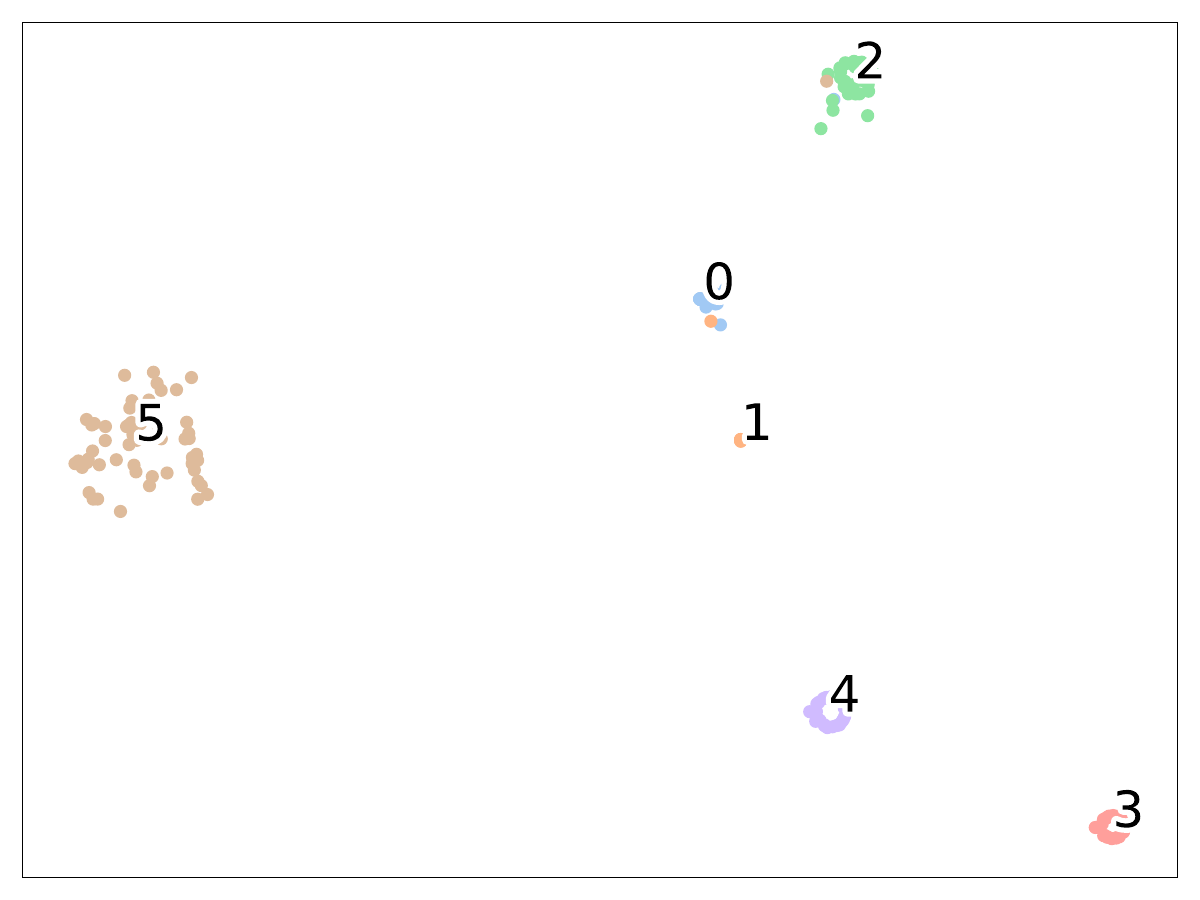}
		\caption{Flow Reps (Default)}
		\label{Flow-level Default}
	\end{subfigure}
        \centering
	\begin{subfigure}{0.49\linewidth}
		\centering
		\includegraphics[width=1.0\linewidth]{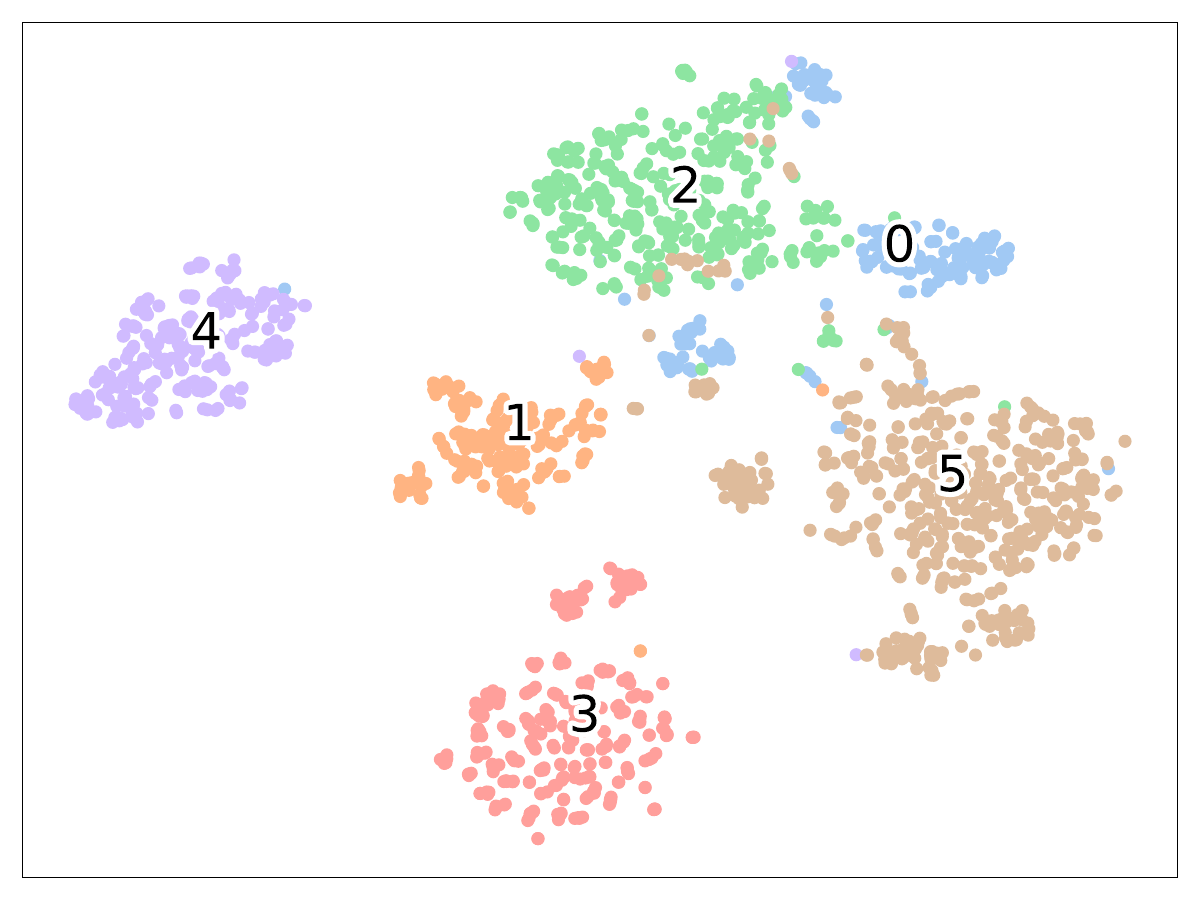}
		\caption{Packet Reps (Default)}
		\label{Packet-level Default}
	\end{subfigure}
        \qquad

	\centering
	\begin{subfigure}{0.49\linewidth}
		\centering
		\includegraphics[width=1.0\linewidth]{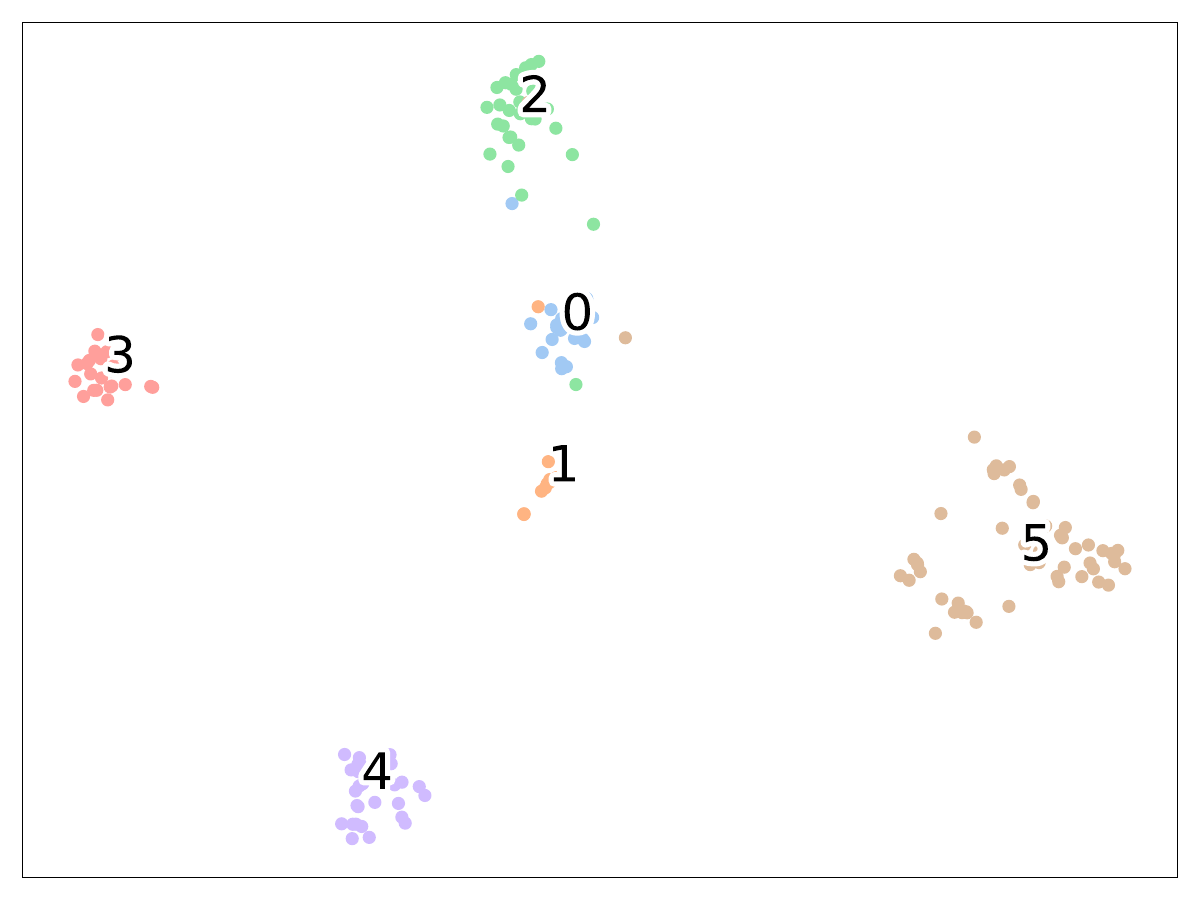}
		\caption{Flow Reps (w/o CL Task)}
		\label{Flow-level w/o CL}
	\end{subfigure}
        \centering
	\begin{subfigure}{0.49\linewidth}
		\centering
		\includegraphics[width=1.0\linewidth]{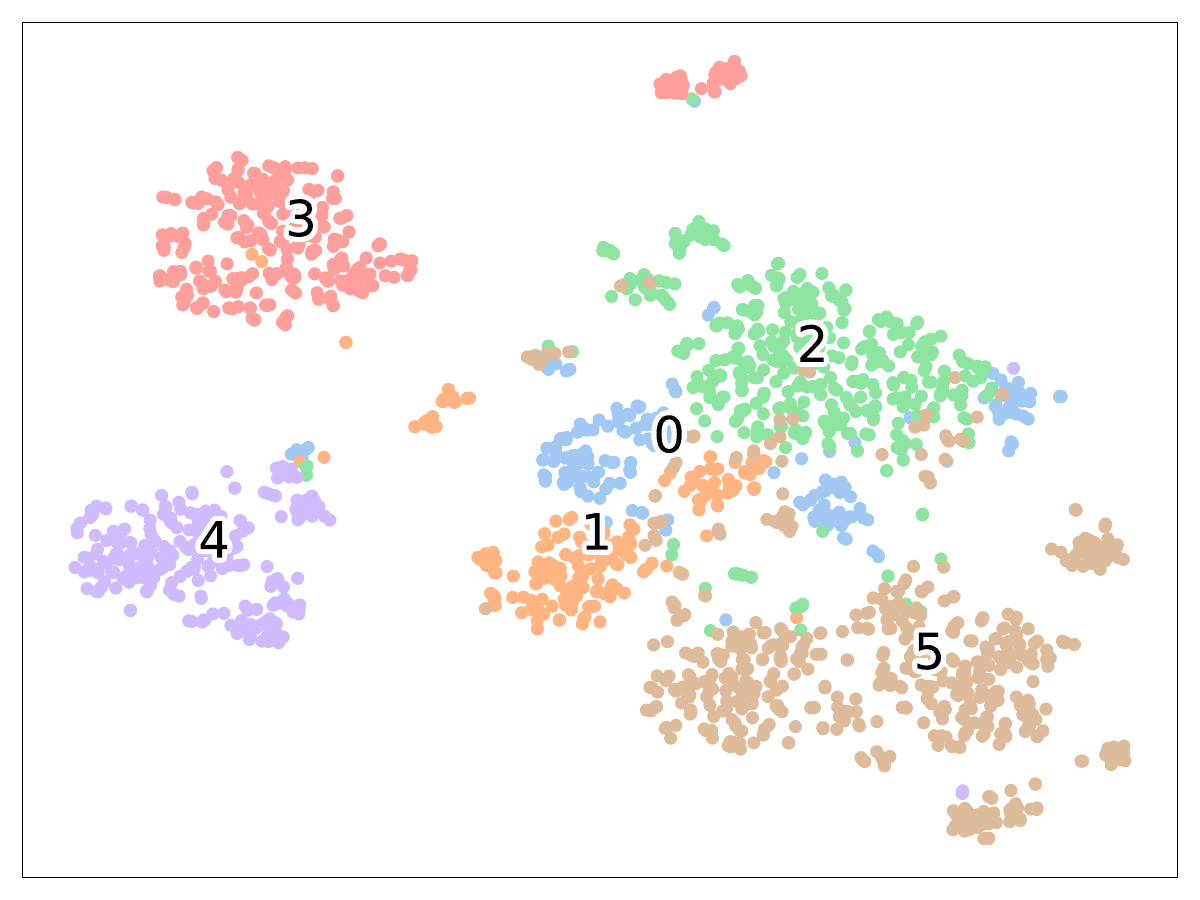}
		\caption{Packet Reps (w/o CL Task)}
		\label{Packet-level w/o CL}
	\end{subfigure}
        \qquad

	\centering
	\begin{subfigure}{0.49\linewidth}
		\centering
		\includegraphics[width=1.0\linewidth]{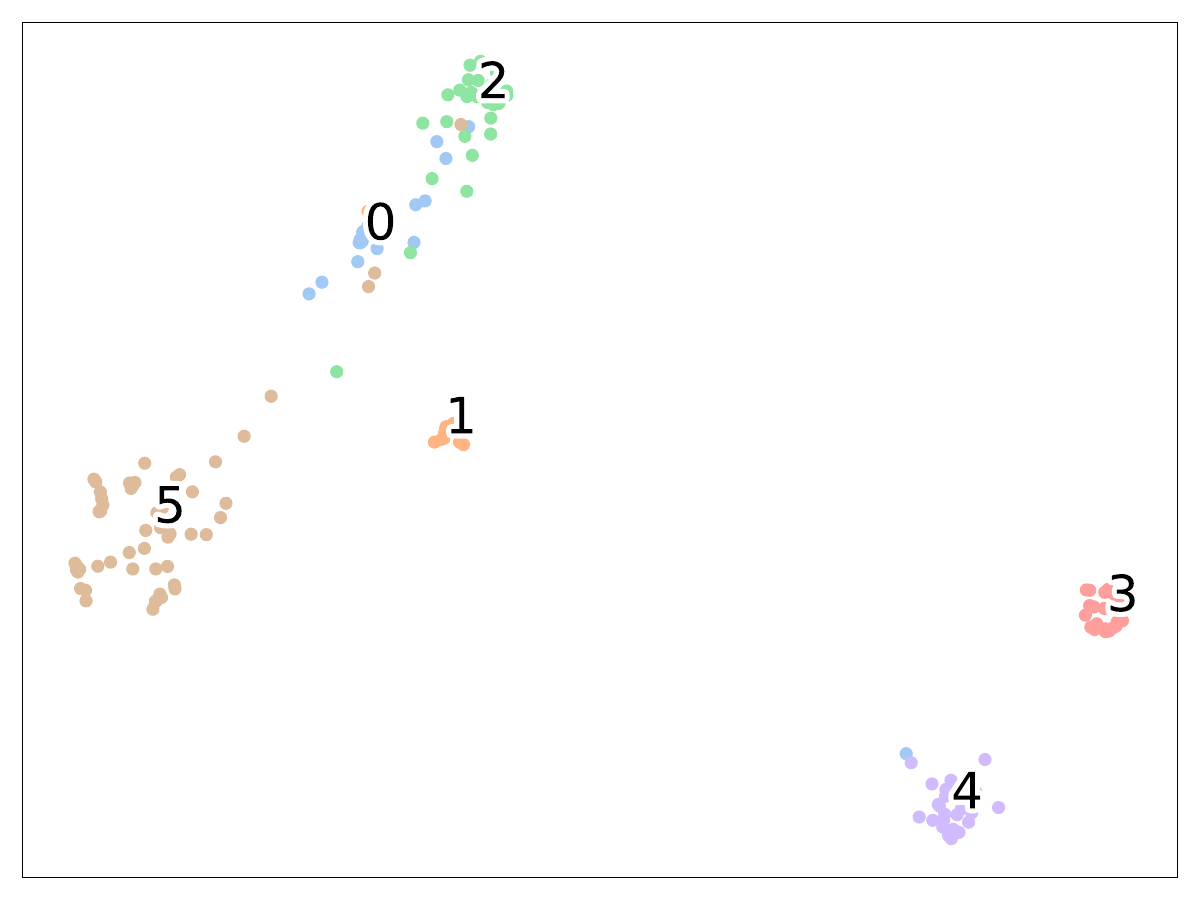}
		\caption{Flow Reps (w/o Packet Tasks)}
		\label{Flow-level w/o Packet-level Task}
	\end{subfigure}
        \centering
	\begin{subfigure}{0.49\linewidth}
		\centering
		\includegraphics[width=1.0\linewidth]{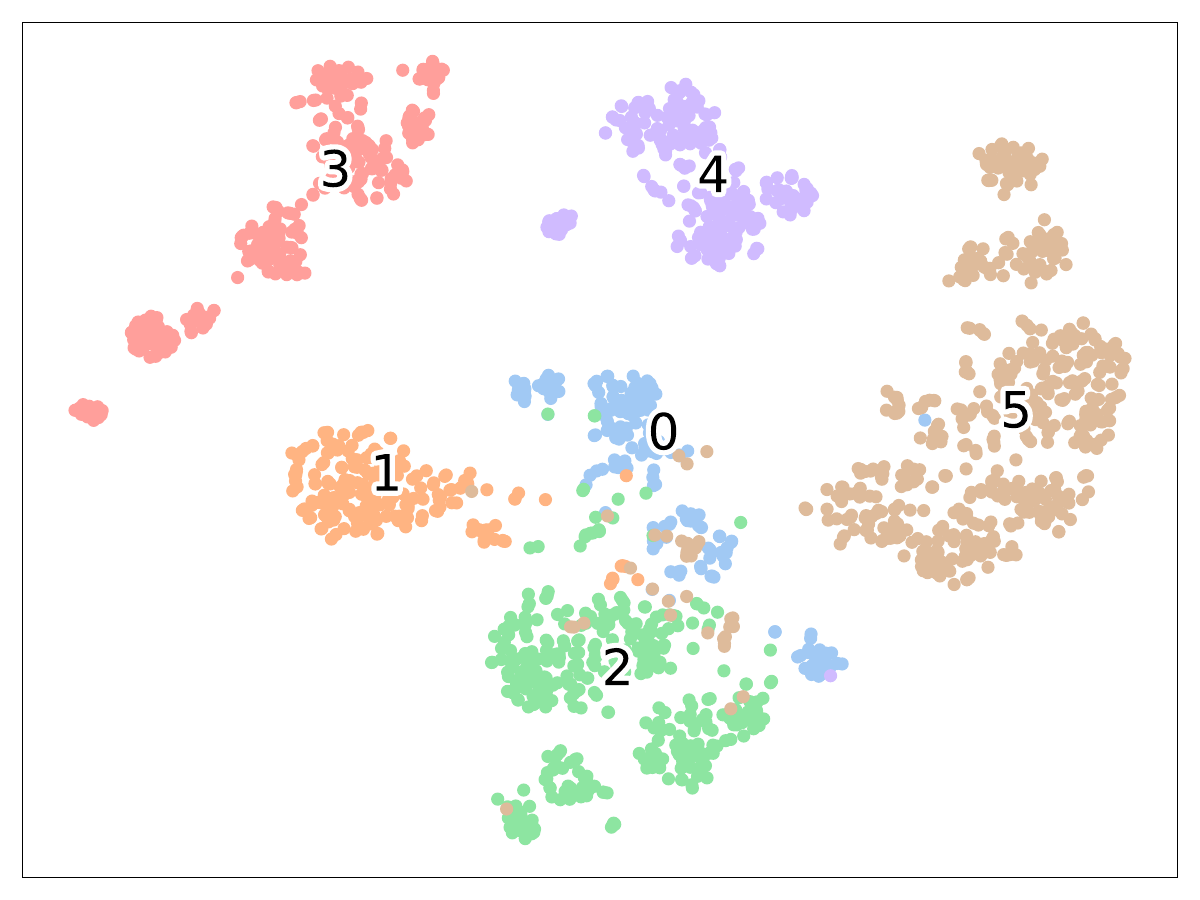}
		\caption{Packet Reps (w/o Flow Tasks)}
		\label{Packet-level w/o Flow-level Task}
	\end{subfigure}
 \vspace{-1ex}
	\caption{t-SNE Visualization w.r.t. the Learned Flow-level and Packet-level Representations (Reps) on the ISCX-VPN Test Dataset (The Digit and Its Coordinate Position Indicate Dataset Category Index and Its Sample Centroid)}
 \vspace{-1ex}
	\label{vpnrepresentation}
\end{figure}

To reveal why contrastive learning can help the model learn more robust representations and how packet-level and flow-level tasks affect each other in joint training, we conduct t-SNE visualization of the packet-level and flow-level representations on the ISCX-VPN test dataset, and the results are shown in Figure \ref{vpnrepresentation}. 

\textbf{Intra-level Relation}. From Figure \ref{Flow-level Default} and \ref{Flow-level w/o CL}, we can find that the flow-level contrastive learning (CL) task obviously brings the flow-level sample representations of the same category closer in the embedding space, thus making the decision boundary between samples of different categories clearer and improving the robustness of the model. 
We can draw similar conclusions from Figure \ref{Packet-level Default} and \ref{Packet-level w/o CL}. 
Since the number of packets is larger than that of flows, we can intuitively see that the packet-level contrastive learning (CL) task widens the average sample distance of different categories in the embedding space. 
However, in Figure \ref{Packet-level w/o CL}, the decision boundary between samples of different categories appears more blurred.

\textbf{Cross-level Relation}. We can obtain some insights from Figure \ref{Flow-level w/o Packet-level Task} and \ref{Packet-level w/o Flow-level Task}. 
To a certain extent, the packet-level task helps the model learn a more robust flow-level representation, which is similar to narrowing the distance between samples of the same category in the embedding space in supervised contrastive learning. 
However, the flow-level task has no obvious positive effect on the packet-level representation. 
More visualization results are in Appendix \ref{sec:appendix_representation}.

\subsection{Computational Cost and Model Parameter Analysis (RQ4)}
\label{sec:exp_compute}

\begin{figure}[H]
\vspace{-1ex}
    \centering
	\begin{subfigure}{0.49\linewidth}
		\centering
		\includegraphics[width=1.0\linewidth]{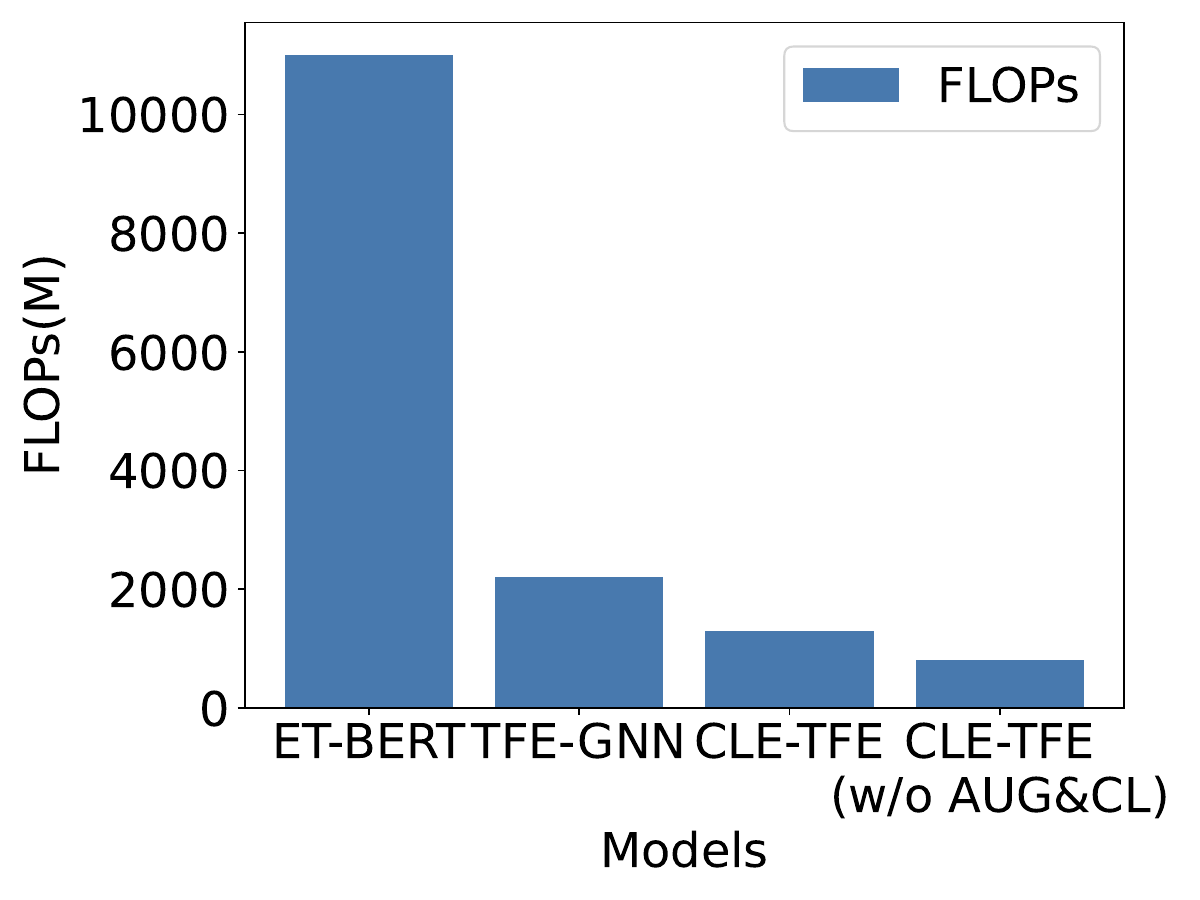}
		\caption{FLOPs(M)}
		\label{FLOPs}
	\end{subfigure}
	\centering
	\begin{subfigure}{0.49\linewidth}
		\centering
		\includegraphics[width=1.0\linewidth]{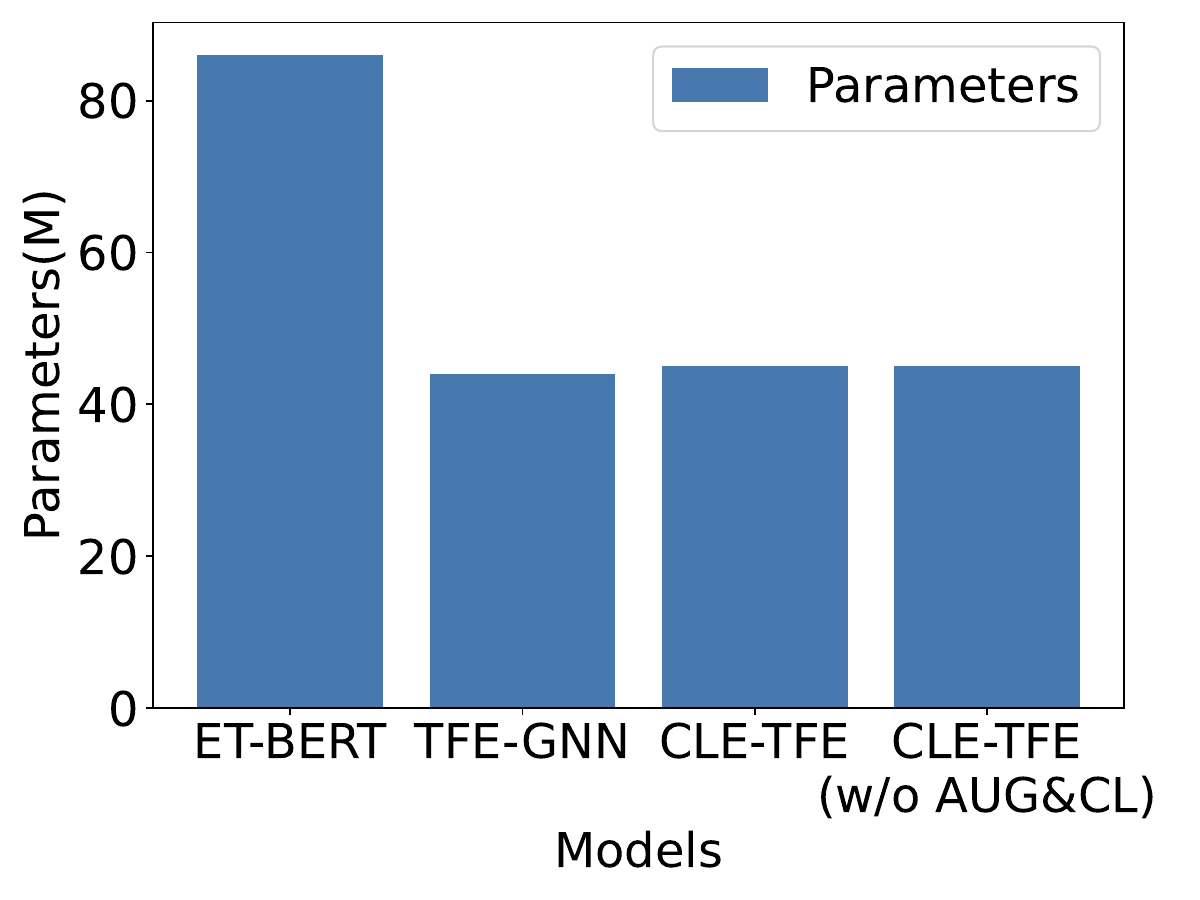}
		\caption{Model Parameters(M)}
		\label{Parameters}
	\end{subfigure}
    \vspace{-2ex}
    \caption{Results of Computational Cost Analysis}
    \vspace{-2ex}
	\label{compute}
\end{figure}

We evaluate the computational costs of CLE-TFE using the floating point operations (FLOPs) and the model parameters and compare CLE-TFE with ET-BERT~\cite{ETBERT}, TFE-GNN~\cite{TFE-GNN}, and CLE-TFE without the augmentation and the calculation of contrastive loss (CLE-TFE w/o AUG\&CL). 
Figure~\ref{compute} shows the results.

Based on Figure~\ref{compute}, we can draw several conclusions. 
(1) ET-BERT, which is a pre-trained model, has the largest FLOPs and parameters, while TFE-GNN is much smaller than ET-BERT. 
(2) Since CLE-TFE only needs a shorter flow length to achieve good results, CLE-TFE has smaller FLOPs than TFE-GNN. 
Furthermore, the parameters of CLE-TFE only increase slightly compared to TFE-GNN (caused by the packet-level classification head), which ensures a relatively small computational cost. 
(3) The augmentation and contrastive loss in CLE-TFE do not add additional parameters, and the additional computational costs brought to CLE-TFE are also very small.

\subsection{Sensitivity Analysis (RQ5)}
\label{sec:exp_sensitivity}

\begin{figure*}[t]
    \centering
	\begin{subfigure}{0.24\linewidth}
		\centering
		\includegraphics[width=1.0\linewidth]{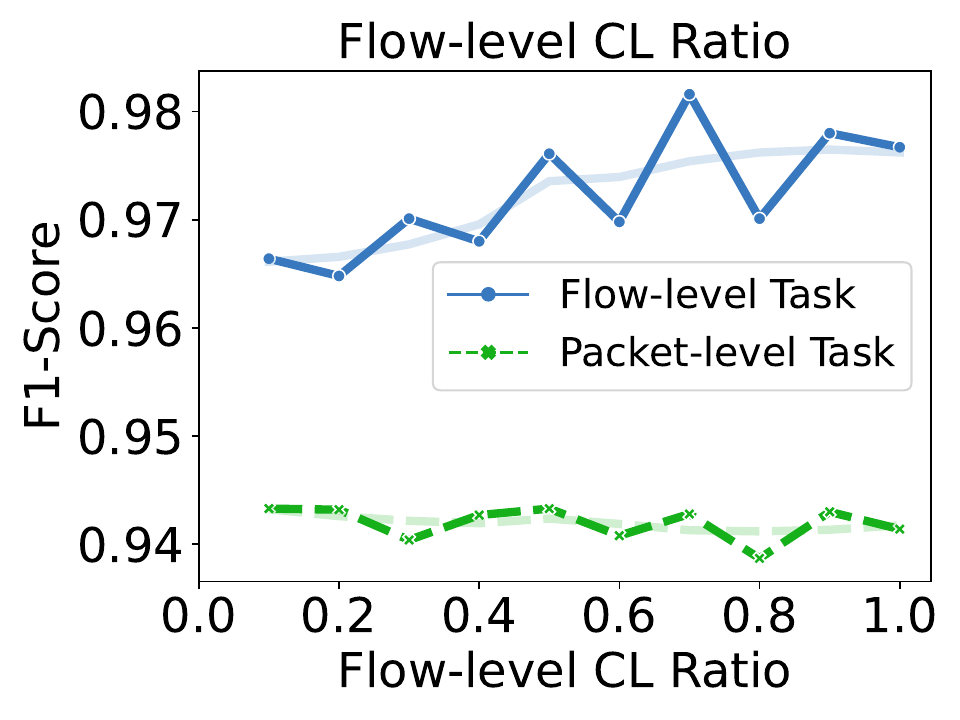}
		\caption{Flow-level CL Ratio $\beta$}
		\label{FCLR}
	\end{subfigure}
	\centering
	\begin{subfigure}{0.24\linewidth}
		\centering
		\includegraphics[width=1.0\linewidth]{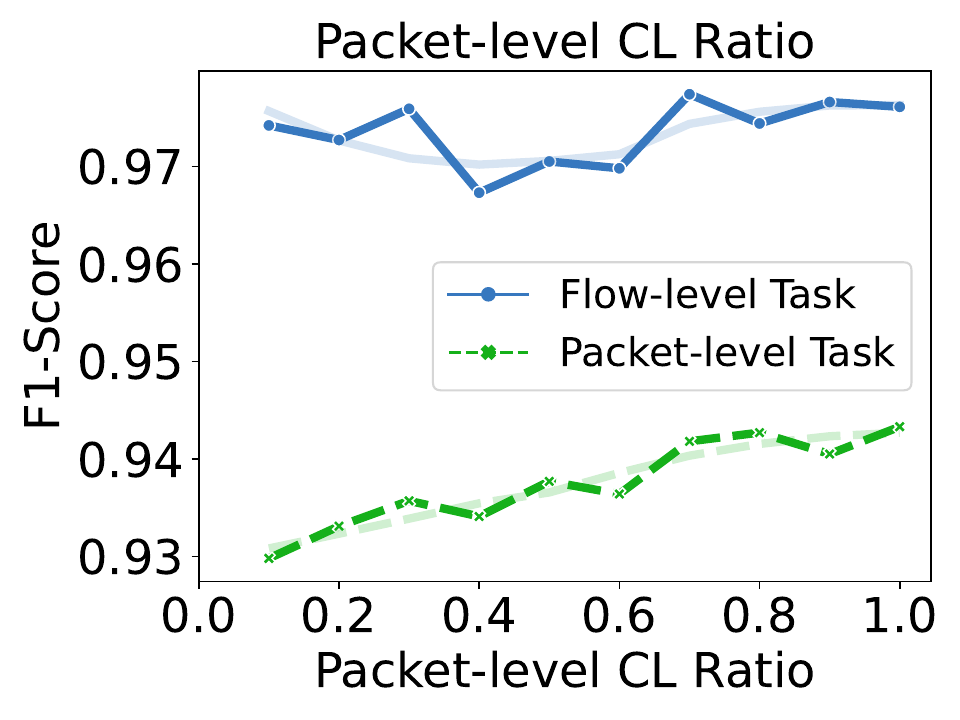}
		\caption{Packet-level CL Ratio $\alpha$}
		\label{PCLR}
	\end{subfigure}
	\centering
	\begin{subfigure}{0.24\linewidth}
		\centering
		\includegraphics[width=1.0\linewidth]{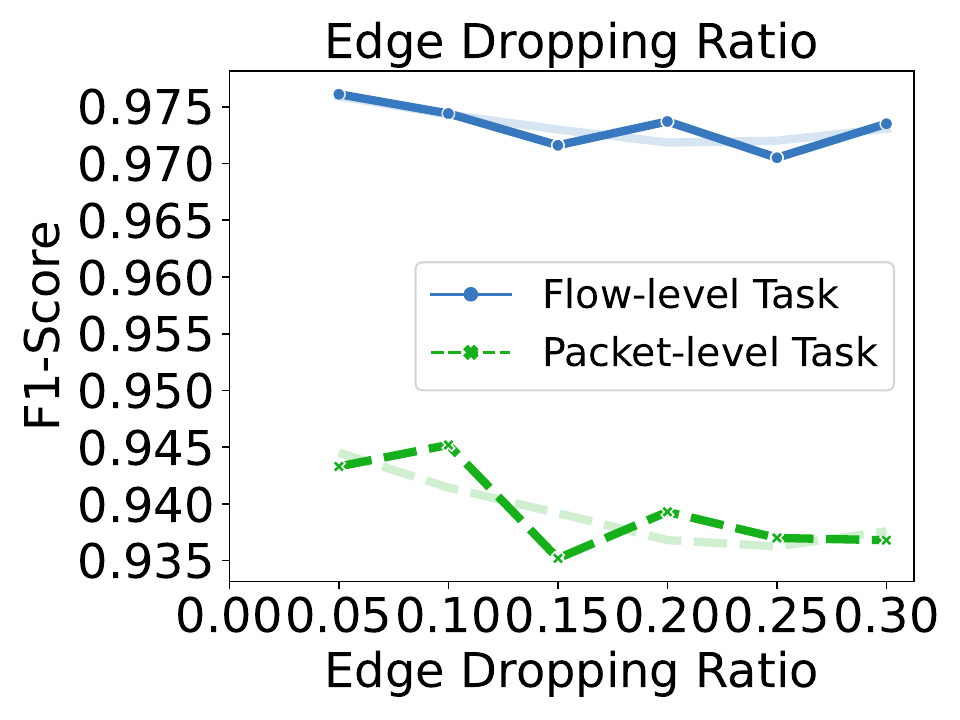}
		\caption{Edge Dropping Ratio $P_{\text{ED}}$}
		\label{EDR}
	\end{subfigure}
        \centering
	\begin{subfigure}{0.24\linewidth}
		\centering
		\includegraphics[width=1.0\linewidth]{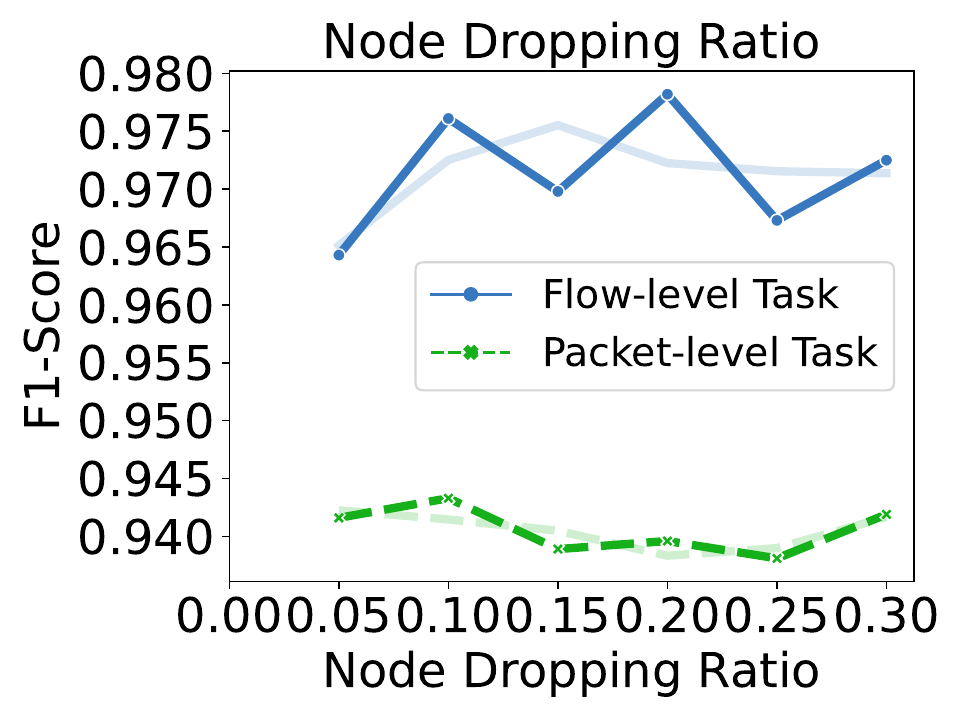}
		\caption{Node Dropping Ratio $P_{\text{ND}}$}
		\label{NDR}
	\end{subfigure}
        \vspace{-1ex}
	\caption{Model Sensitivity Analysis w.r.t. Flow-level CL Ratio $\beta$, Packet-level CL Ratio $\alpha$, Edge Dropping Ratio $P_{\text{ED}}$, and Node Dropping Ratio $P_{\text{ND}}$ on the ISCX-VPN Dataset (The shallow lines are attained by smoothing the dark plotted lines)}
	\vspace{-1ex}
	\label{sensitivity}
\end{figure*}

To investigate the influence of hyper-parameters on model performance, we conduct sensitivity analysis w.r.t. flow-level contrastive learning (cl) ratio $\beta$, packet-level cl ratio $\alpha$, edge dropping ratio $P_{\text{ED}}$, and node dropping ratio $P_{\text{ND}}$ on the ISCX-VPN dataset, and the results are shown in Figure \ref{sensitivity}. 

\textbf{Flow-level CL Ratio $\beta$}. From Figure \ref{FCLR}, increasing the flow-level cl ratio can enhance the performance of the flow-level task but has little effect on the packet-level task. 
\textbf{Packet-level CL Ratio $\alpha$}. From Figure \ref{PCLR}, increasing the packet-level cl ratio can make the packet-level task perform better, but it will also fluctuate the performance of the flow-level task. 
\textbf{Edge Dropping Ratio $P_{\text{ED}}$}. From Figure \ref{EDR}, we can conclude that a small edge dropping ratio is enough to make the model achieve superior results, while a large one may introduce too much noise and cause model performance degradation. 
\textbf{Node Dropping Ratio $P_{\text{ND}}$}. From Figure \ref{NDR}, we can find that node dropping ratio has a greater impact on the model performance than edge dropping ratio because deleting a node will also delete the edges connected to it. 
But generally speaking, a small ratio is preferred to reach a better f1-score. 
More sensitivity analysis results are in Appendix \ref{sec:appendix_sensitivity}.

\section{Related Work}
\label{sec:related_work}

\noindent \textbf{Flow-level Traffic Classification Methods.} 
Many flow-level traffic classification methods have been proposed.

$\bullet$ \noindent\textit{Statistical Feature Based Methods.} 
Many methods use statistical features to represent packet properties and utilize traditional machine learning models for classification. 
AppScanner~\cite{AppScanner} extracts features from traffic flows based on bidirectional flow characteristics. 
CUMUL~\cite{CUMUL} uses cumulative packet length as its feature, while GRAIN~\cite{GRAIN} uses payload length. 
ETC-PS~\cite{ETC-PS} strengthens packet length sequences by applying the path signature theory, and Liu \emph{et al.}~\cite{FAAR} uses wavelet decomposition to exploit them. 
Hierarchical clustering is also leveraged for feature extraction by Conti \emph{et al.}~\cite{Conti}. 

$\bullet$ \noindent\textit{Fingerprinting Matching Based Methods.} 
Fingerprinting denotes traffic characteristics and is also used in traffic identification. 
FlowPrint~\cite{FlowPrint} generates traffic fingerprints by creating correlation graphs that compute activity values between destination IPs. 
K-FP~\cite{KFP} uses the random forest to construct fingerprints and identifies unknown samples through k-nearest neighbor matching.
These two methods cannot encode an independent packet into a high-dimensional vector, thus failing in packet-level classification tasks. 

$\bullet$ \noindent\textit{Deep Learning Based Methods.} 
Deep learning has demonstrated powerful learning abilities, and many traffic classification methods are based on it. 
MVML~\cite{MVML} extracts local and global features using packet length and time delay sequences, then leverages multilayer perceptions for traffic identification. 
EDC~\cite{EDC} utilizes packet header properties (e.g., protocol types, packet length) to serve as inputs for MLPs. 
RBRN~\cite{RBRN}, DF~\cite{DF}, and FS-Net~\cite{FSNet} all use statistical feature sequences (e.g., packet length sequences) as inputs for convolutional neural networks (CNNs) or recurrent neural networks (RNNs). 
There are also some methods using raw bytes as features. 
FFB~\cite{FFB} utilizes raw bytes and packet length sequences as inputs for CNNs and RNNs. 
EBSNN~\cite{EBSNN} combines RNNs with the attention mechanism to process header and payload byte segments. 
ET-BERT~\cite{ETBERT} conducts pre-training tasks on large-scale traffic datasets to learn a powerful raw byte representation, which is time-consuming and expensive. 
Graph neural networks (GNNs) are another model that can be used for traffic classification tasks. 
GraphDApp~\cite{GraphDApp} builds traffic interaction graphs from traffic bursts and uses GNNs for representation learning. 
Huoh \emph{et al.}~\cite{ECD-GNN} generate edges based on the chronological order of packets in a flow. 
TFE-GNN~\cite{TFE-GNN} employs point-wise mutual information~\cite{PMI} to construct byte-level traffic graphs and designs a traffic graph encoder for feature extraction. 
These methods conduct the traffic classification task at the flow level, which cannot benefit from the cross-level relation with the packet-level classification task (Sec~ \ref{sec:exp_ablation}, \ref{sec:exp_representation}).

\noindent \textbf{Packet-level Traffic Classification Methods.} 
KISS~\cite{KISS} extracts statistical signatures of payload to train SVM for UDP packet classification. 
HEDGE~\cite{HEDGE} is a threshold-based method using randomness tests of payload to classify a single packet without accessing the entire stream. 
Securitas~\cite{Securitas} generates n-grams for raw bytes and utilizes Latent Dirichlet Allocation (LDA) to form protocol keywords as features, followed by SVM, C4.5 decision tree, or Bayes network for packet classification. 
2D-CNN~\cite{2DCNN} and 3D-CNN~\cite{3DCNN} treat packet bytes as pixel values and convert them into images, which are further fed into 2D-CNNs and 3D-CNNs for packet classification. 
DP~\cite{DeepPacket} leverages CNNs and autoencoders to extract byte features. 
BLJAN~\cite{BLJAN} explores the correlation between packet bytes and their labels and encodes them into a joint embedding space to classify packets. 
EBSNN~\cite{EBSNN} and ET-BERT~\cite{ETBERT} can also perform packet classification. But, they still require two independent training or fine-tuning to conduct flow-level and packet-level tasks, which are computationally expensive. 
PacRep~\cite{PacRep} leverages triplet loss~\cite{tripletloss} without data augmentation and jointly optimizes multiple packet-level tasks to learn a better packet representation. 
However, it is still based on a single packet level, and the transformer-based~\cite{Self-Attention} pre-trained encoder is very computationally intensive.

\section{Conclusion and Future Work}
\label{sec:conclusion}
In this paper, we propose a simple yet effective model named CLE-TFE that uses supervised contrastive learning to obtain more robust packet-level and flow-level representations. 
In particular, we perform graph data augmentation on the byte-level traffic graph of TFE-GNN to uncover fine-grained semantic-invariant representations between bytes through contrastive learning. 
Through cross-level multi-task learning, we can perform the packet-level and flow-level classification tasks in the same model with one training. 
The experimental results show that CLE-TFE reaches the overall best performance on the two tasks with slight computational costs.

In the future, we will continue to pursue research on the following several topics: 
\textbf{(1) Learnable Graph Augmentation}. The graph augmentation is non-trainable, which may give sub-optimal results. 
\textbf{(2) Hard Negative Sample Mining in Contrastive Learning}. Although we utilize supervised contrastive learning to increase the positive sample pairs, some hard negative samples should be optimized more sufficiently for better performance.

\newpage

\bibliographystyle{ACM-Reference-Format}
\bibliography{ref}

\newpage
\appendix
\section{Threat Model and Assumptions}
\label{sec:appendix_threat}

We briefly describe the threat model and assumptions in this section. 

Normal users employ mobile apps to communicate with remote servers. The attacker is a passive observer (i.e., he cannot decrypt or modify packets). The attacker captures the packets of the target apps by compromising the device or sniffing the network link. Then, the attacker analyzes the captured packets to infer the behaviors of normal users.

\section{More Ablation Study Results}
\label{sec:appendix_ablation}

In this section, we give more ablation study results on the ISCX-NonVPN, ISCX-Tor, and ISCX-NonTor datasets. The results are shown in Table~\ref{tab:Ablation_study_nonvpn}, \ref{tab:Ablation_study_tor}, and \ref{tab:Ablation_study_nontor}, respectively.

\begin{table*}[t]
  \footnotesize
  \caption{Ablation Study of CLE-TFE on the ISCX-NonVPN Dataset}
  \label{tab:Ablation_study_nonvpn}
  \begin{tabular}{c|cccc|cccc|cccc}
    \toprule
    Tasks & \multicolumn{4}{c|}{Flow-level Classification Task} & \multicolumn{4}{c|}{Packet-level Classification Task} & \multicolumn{4}{c}{Average on the Two Tasks} \\
    \midrule
    Variants & AC & PR & RC & F1 & AC & PR & RC & F1 & AC & PR & RC & F1 \\
    \midrule
    w/o Flow-level Classification Loss
    & - & - & - & - 
    & 0.8855 & \textbf{0.9046} & 0.9024 & 0.9025
    & 0.4428 & 0.4523 & 0.4512 & 0.4513 \\
    w/o Flow-level Contrastive Loss
    & 0.9192 & 0.9306 & 0.9301 & 0.9301 
    & \textbf{0.8858} & 0.9033 & \textbf{0.9033} & 0.9025
    & 0.9025 & 0.9170 & 0.9167 & 0.9163  \\
    w/o Flow-level Classification \& Contrastive Loss
    & - & - & - & - 
    & 0.8855 & 0.9045 & 0.9032 & \textbf{0.9027}
    & 0.4428 & 0.4523 & 0.4516 & 0.4514 \\
    \midrule
    w/o Packet-level Classification Loss
    & 0.9232 & 0.9365 & 0.9284 & 0.9321 
    & - & - & - & - 
    & 0.4616 & 0.4683 & 0.4642 & 0.4661 \\
    w/o Packet-level Contrastive Loss
    & 0.9136 & 0.9248 & 0.9240 & 0.9241 
    & 0.8744 & 0.8929 & 0.8923 & 0.8911 
    & 0.8940 & 0.9089 & 0.9082 & 0.9076 \\
    w/o Packet-level Classification \& Contrastive Loss
    & 0.9061 & 0.9216 & 0.9122 & 0.9161 
    & - & - & - & - 
    & 0.4531 & 0.4608 & 0.4561 & 0.4581 \\
    \midrule
    w/o Packet-level Header Graph Aug
    & 0.9207 & 0.9331 & 0.9302 & 0.9313 
    & 0.8805 & 0.8987 & 0.8983 & 0.8976 
    & 0.9006 & 0.9159 & 0.9143 & 0.9145 \\
    w/o Packet-level Payload Graph Aug
    & 0.9237 & 0.9342 & 0.9336 & 0.9336 
    & 0.8838 & 0.9014 & 0.9021 & 0.9012 
    & 0.9038 & 0.9178 & 0.9179 & 0.9174 \\
    w/o Packet-level Header \& Payload Graph Aug
    & 0.9222 & 0.9344 & 0.9284 & 0.9311 
    & 0.8797 & 0.8985 & 0.8971 & 0.8966 
    & 0.9010 & 0.9165 & 0.9128 & 0.9139 \\
    \midrule
    w/ Unsupervised Contrastive Loss (No Label Used)
    & 0.8096 & 0.8233 & 0.8119 & 0.8137 
    & 0.8396 & 0.8540 & 0.8603 & 0.8555 
    & 0.8246 & 0.8387 & 0.8361 & 0.8346 \\
    \midrule
    \textbf{CLE-TFE}
    & \textbf{0.9286} & \textbf{0.9396} & \textbf{0.9391} & \textbf{0.9389} 
    & 0.8853 & 0.9039 & 0.9033 & 0.9027 
    & \textbf{0.9070} & \textbf{0.9218} & \textbf{0.9212} & \textbf{0.9208} \\
    \bottomrule
  \end{tabular}
\end{table*}

\begin{table*}[t]
  \footnotesize
  \caption{Ablation Study of CLE-TFE on the ISCX-Tor Dataset}
  \label{tab:Ablation_study_tor}
  \begin{tabular}{c|cccc|cccc|cccc}
    \toprule
    Tasks & \multicolumn{4}{c|}{Flow-level Classification Task} & \multicolumn{4}{c|}{Packet-level Classification Task} & \multicolumn{4}{c}{Average on the Two Tasks} \\
    \midrule
    Variants & AC & PR & RC & F1 & AC & PR & RC & F1 & AC & PR & RC & F1 \\
    \midrule
    w/o Flow-level Classification Loss
    & - & - & - & - 
    & 0.9994 & 0.9993 & 0.9997 & 0.9995
    & 0.4997 & 0.4997 & 0.4999 & 0.4998 \\
    w/o Flow-level Contrastive Loss
    & 0.9989 & 0.9983 & 0.9994 & 0.9988 
    & 0.9990 & 0.9987 & 0.9980 & 0.9983
    & 0.9990 & 0.9985 & 0.9987 & 0.9986  \\
    w/o Flow-level Classification \& Contrastive Loss
    & - & - & - & - 
    & 0.9985 & 0.9988 & 0.9992 & 0.9990
    & 0.4993 & 0.4994 & 0.4996 & 0.4995 \\
    \midrule
    w/o Packet-level Classification Loss
    & \textbf{1.0000} & \textbf{1.0000} & \textbf{1.0000} & \textbf{1.0000} 
    & - & - & - & - 
    & 0.5000 & 0.5000 & 0.5000 & 0.5000 \\
    w/o Packet-level Contrastive Loss
    & 0.9989 & 0.9964 & 0.9994 & 0.9978 
    & 0.9971 & 0.9957 & 0.9980 & 0.9968 
    & 0.9980 & 0.9961 & 0.9987 & 0.9973 \\
    w/o Packet-level Classification \& Contrastive Loss
    & 0.9954 & 0.9976 & 0.9976 & 0.9975 
    & - & - & - & - 
    & 0.4977 & 0.4988 & 0.4988 & 0.4988 \\
    \midrule
    w/o Packet-level Header Graph Aug
    & \textbf{1.0000} & \textbf{1.0000} & \textbf{1.0000} & \textbf{1.0000} 
    & 0.9993 & 0.9987 & 0.9996 & 0.9992 
    & 0.9997 & 0.9994 & 0.9998 & 0.9996 \\
    w/o Packet-level Payload Graph Aug
    & 0.9989 & 0.9964 & 0.9994 & 0.9978 
    & 0.9991 & 0.9986 & 0.9996 & 0.9990 
    & 0.9990 & 0.9975 & 0.9995 & 0.9984 \\
    w/o Packet-level Header \& Payload Graph Aug
    & \textbf{1.0000} & \textbf{1.0000} & \textbf{1.0000} & \textbf{1.0000} 
    & \textbf{0.9996} & 0.9992 & \textbf{0.9998} & 0.9995 
    & \textbf{0.9998} & 0.9996 & \textbf{0.9999} & 0.9998 \\
    \midrule
    w/ Unsupervised Contrastive Loss (No Label Used)
    & 0.9337 & 0.9467 & 0.9459 & 0.9410 
    & 0.9791 & 0.9800 & 0.9817 & 0.9803 
    & 0.9564 & 0.9634 & 0.9638 & 0.9607 \\
    \midrule
    \textbf{CLE-TFE}
    & \textbf{1.0000} & \textbf{1.0000} & \textbf{1.0000} & \textbf{1.0000} 
    & \textbf{0.9996} & \textbf{0.9997} & \textbf{0.9998} & \textbf{0.9997} 
    & \textbf{0.9998} & \textbf{0.9999} & \textbf{0.9999} & \textbf{0.9999} \\
    \bottomrule
  \end{tabular}
\end{table*}

\begin{table*}[t]
  \footnotesize
  \caption{Ablation Study of CLE-TFE on the ISCX-NonTor Dataset}
  \label{tab:Ablation_study_nontor}
  \begin{tabular}{c|cccc|cccc|cccc}
    \toprule
    Tasks & \multicolumn{4}{c|}{Flow-level Classification Task} & \multicolumn{4}{c|}{Packet-level Classification Task} & \multicolumn{4}{c}{Average on the Two Tasks} \\
    \midrule
    Variants & AC & PR & RC & F1 & AC & PR & RC & F1 & AC & PR & RC & F1 \\
    \midrule
    w/o Flow-level Classification Loss
    & - & - & - & - 
    & \textbf{0.9259} & 0.8863 & \textbf{0.8562} & \textbf{0.8688}
    & 0.4630 & 0.4432 & 0.4281 & 0.4344 \\
    w/o Flow-level Contrastive Loss
    & 0.9505 & 0.8876 & 0.8844 & 0.8844 
    & 0.9203 & 0.8776 & 0.8443 & 0.8576
    & 0.9354 & 0.8826 & 0.8644 & 0.8710  \\
    w/o Flow-level Classification \& Contrastive Loss
    & - & - & - & - 
    & 0.9228 & \textbf{0.8873} & 0.8557 & 0.8685
    & 0.4614 & 0.4437 & 0.4279 & 0.4343 \\
    \midrule
    w/o Packet-level Classification Loss
    & 0.9504 & 0.8851 & 0.8761 & 0.8793 
    & - & - & - & - 
    & 0.4752 & 0.4426 & 0.4381 & 0.4397 \\
    w/o Packet-level Contrastive Loss
    & 0.9531 & \textbf{0.9066} & 0.8924 & 0.8989 
    & 0.9164 & 0.8733 & 0.8389 & 0.8527 
    & 0.9348 & 0.8900 & 0.8657 & 0.8758 \\
    w/o Packet-level Classification \& Contrastive Loss
    & 0.9497 & 0.8843 & 0.8772 & 0.8797 
    & - & - & - & - 
    & 0.4749 & 0.4422 & 0.4386 & 0.4399 \\
    \midrule
    w/o Packet-level Header Graph Aug
    & 0.9509 & 0.8923 & 0.8812 & 0.8857 
    & 0.9233 & 0.8810 & 0.8496 & 0.8625 
    & 0.9371 & 0.8867 & 0.8654 & 0.8741 \\
    w/o Packet-level Payload Graph Aug
    & 0.9492 & 0.8856 & 0.8869 & 0.8849 
    & 0.9230 & 0.8746 & 0.8512 & 0.8613 
    & 0.9361 & 0.8801 & 0.8691 & 0.8731 \\
    w/o Packet-level Header \& Payload Graph Aug
    & 0.9507 & 0.8987 & 0.8884 & 0.8907 
    & 0.9222 & 0.8748 & 0.8512 & 0.8612 
    & 0.9365 & 0.8868 & 0.8698 & 0.8760 \\
    \midrule
    w/ Unsupervised Contrastive Loss (No Label Used)
    & 0.9148 & 0.8765 & 0.7807 & 0.8124 
    & 0.8710 & 0.8240 & 0.7817 & 0.7977 
    & 0.8929 & 0.8503 & 0.7812 & 0.8051 \\
    \midrule
    \textbf{CLE-TFE}
    & \textbf{0.9554} & 0.9009 & \textbf{0.9019} & \textbf{0.8994} 
    & 0.9240 & 0.8795 & 0.8544 & 0.8649 
    & \textbf{0.9397} & \textbf{0.8902} & \textbf{0.8782} & \textbf{0.8822} \\
    \bottomrule
  \end{tabular}
\end{table*}

\section{More Sensitivity Analysis Results}
\label{sec:appendix_sensitivity}

\begin{figure*}[t]
    \centering
	\begin{subfigure}{0.24\linewidth}
		\centering
		\includegraphics[width=1.0\linewidth]{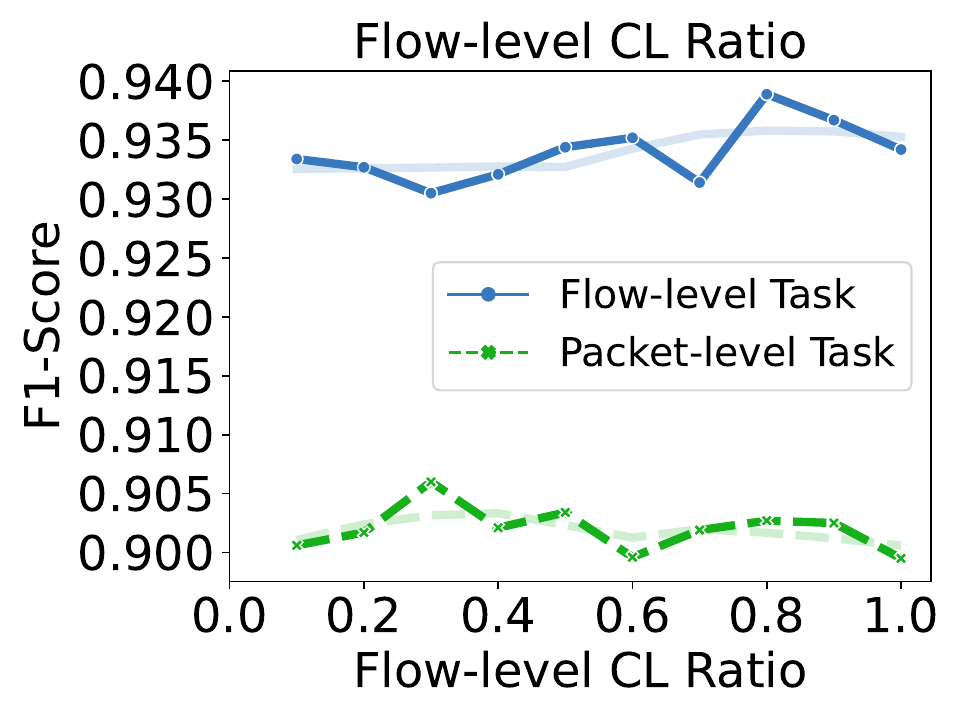}
		\caption{Flow-level CL Ratio $\beta$}
		\label{NonVPNFCLR}
	\end{subfigure}
	\centering
	\begin{subfigure}{0.24\linewidth}
		\centering
		\includegraphics[width=1.0\linewidth]{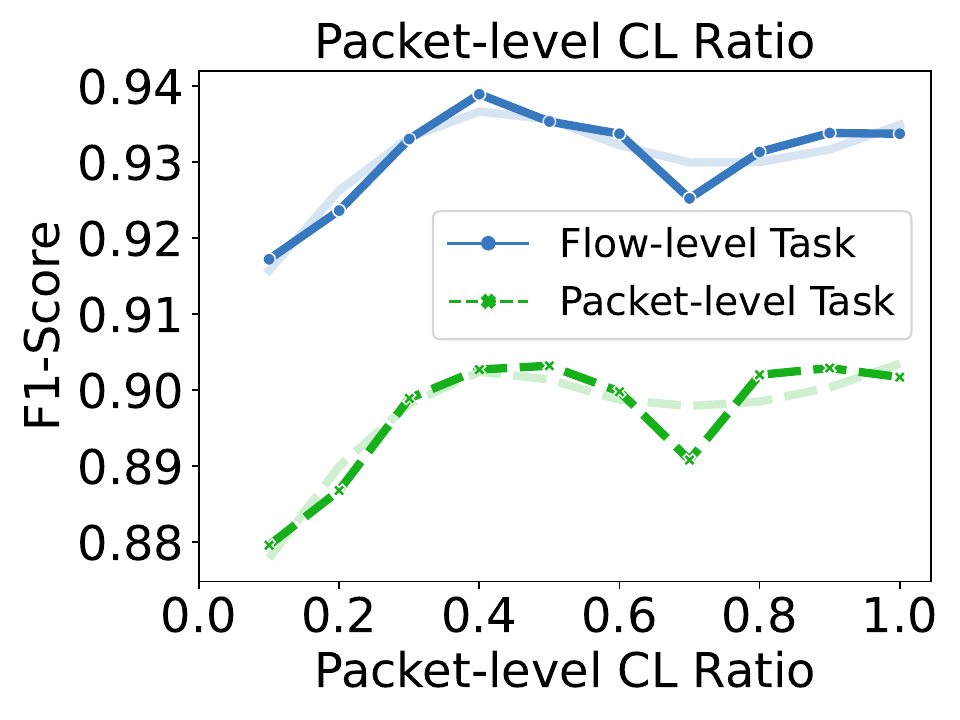}
		\caption{Packet-level CL Ratio $\alpha$}
		\label{NonVPNPCLR}
	\end{subfigure}
	\centering
	\begin{subfigure}{0.24\linewidth}
		\centering
		\includegraphics[width=1.0\linewidth]{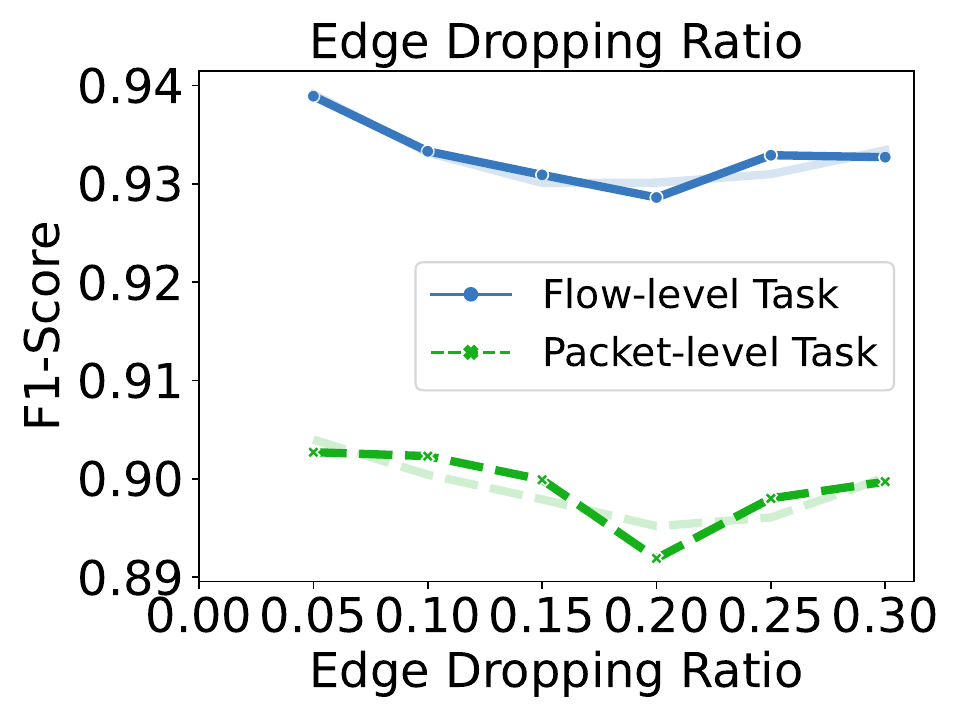}
		\caption{Edge Dropping Ratio $P_{\text{ED}}$}
		\label{NonVPNEDR}
	\end{subfigure}
        \centering
	\begin{subfigure}{0.24\linewidth}
		\centering
		\includegraphics[width=1.0\linewidth]{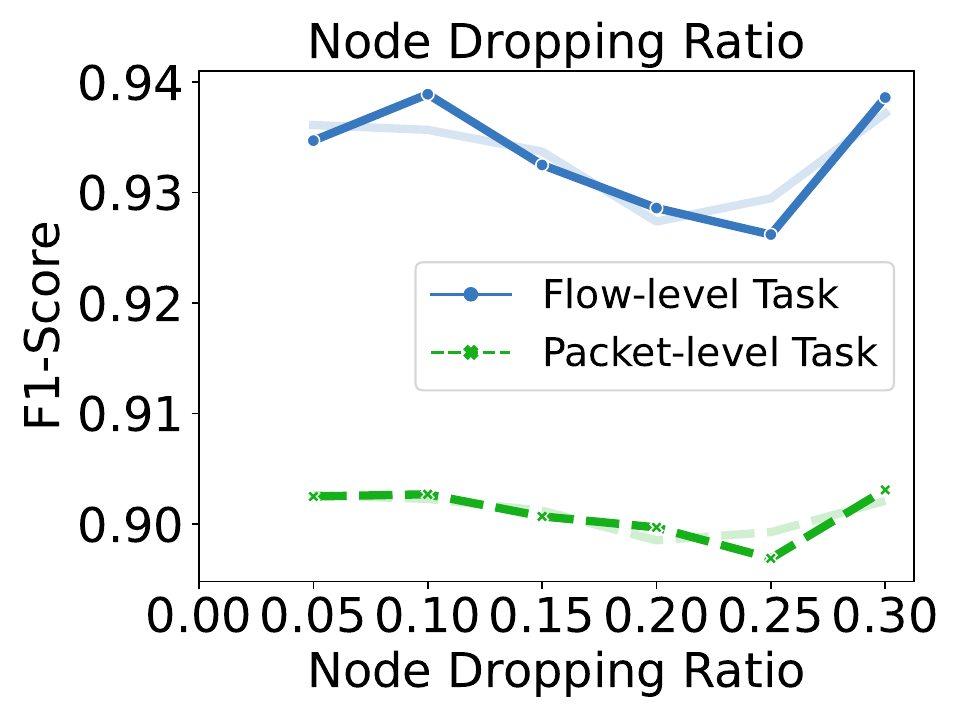}
		\caption{Node Dropping Ratio $P_{\text{ND}}$}
		\label{NonVPNNDR}
	\end{subfigure}
	\caption{Model Sensitivity Analysis w.r.t. Flow-level CL Ratio $\beta$, Packet-level CL Ratio $\alpha$, Edge Dropping Ratio $P_{\text{ED}}$, and Node Dropping Ratio $P_{\text{ND}}$ on the ISCX-NonVPN Dataset (The shallow lines are attained by smoothing the dark plotted lines)}
	\label{NonVPNsensitivity}
\end{figure*}

\begin{figure*}[t]
    \centering
	\begin{subfigure}{0.24\linewidth}
		\centering
		\includegraphics[width=1.0\linewidth]{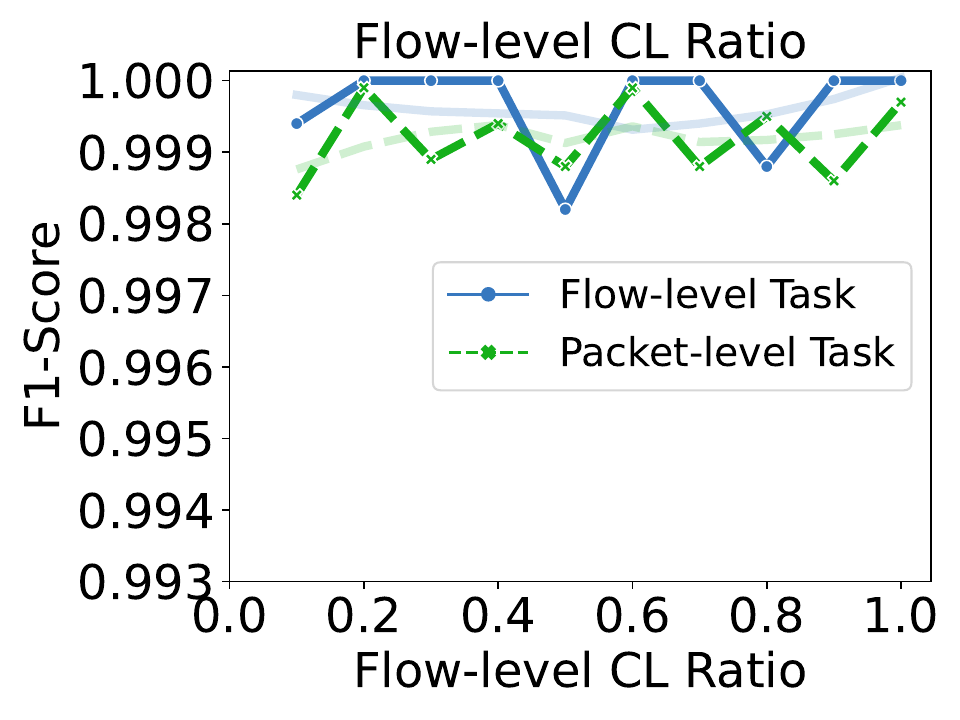}
		\caption{Flow-level CL Ratio $\beta$}
		\label{TorFCLR}
	\end{subfigure}
	\centering
	\begin{subfigure}{0.24\linewidth}
		\centering
		\includegraphics[width=1.0\linewidth]{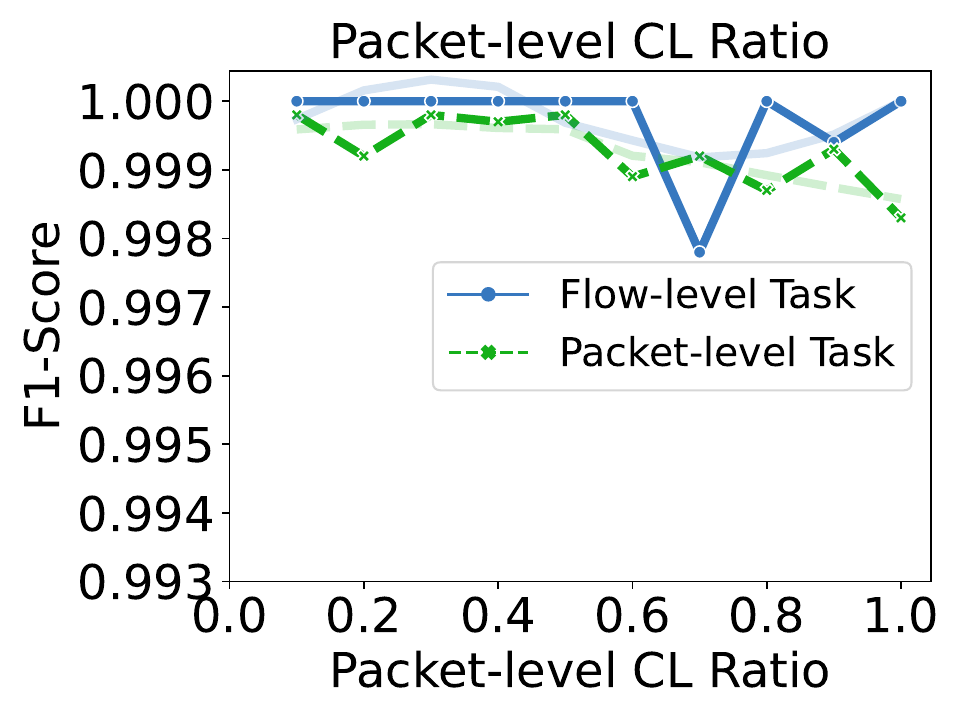}
		\caption{Packet-level CL Ratio $\alpha$}
		\label{TorPCLR}
	\end{subfigure}
	\centering
	\begin{subfigure}{0.24\linewidth}
		\centering
		\includegraphics[width=1.0\linewidth]{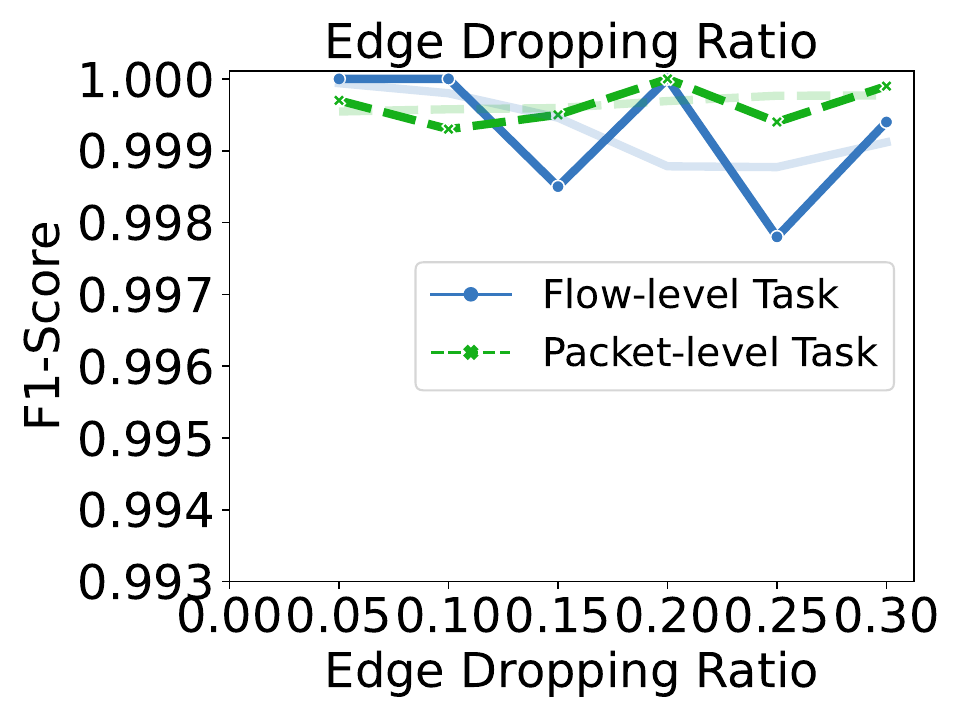}
		\caption{Edge Dropping Ratio $P_{\text{ED}}$}
		\label{TorEDR}
	\end{subfigure}
        \centering
	\begin{subfigure}{0.24\linewidth}
		\centering
		\includegraphics[width=1.0\linewidth]{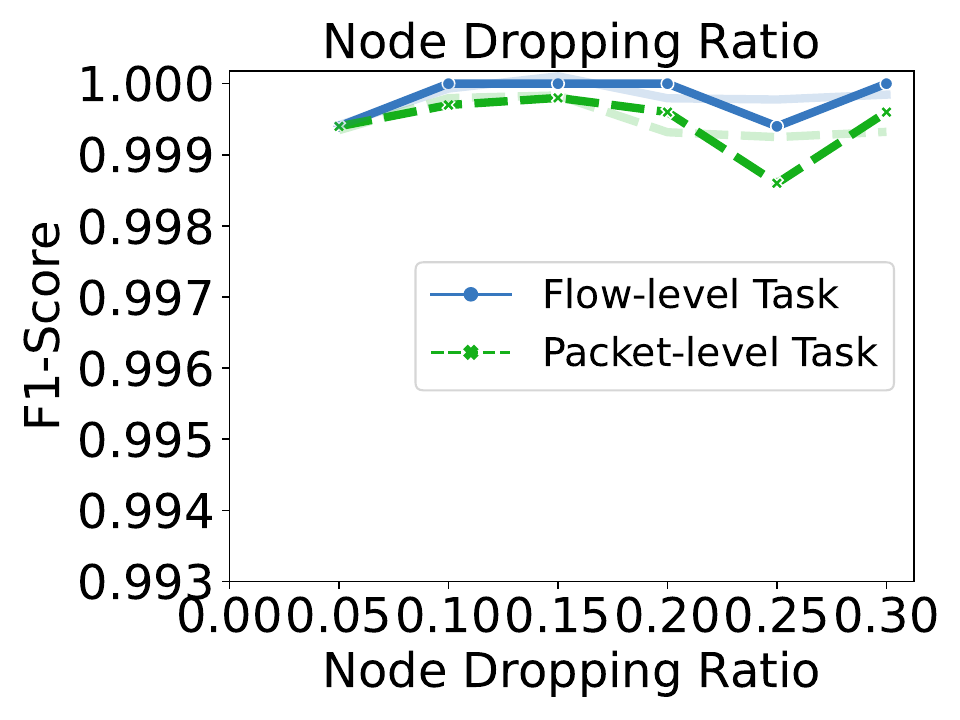}
		\caption{Node Dropping Ratio $P_{\text{ND}}$}
		\label{TorNDR}
	\end{subfigure}
	\caption{Model Sensitivity Analysis w.r.t. Flow-level CL Ratio $\beta$, Packet-level CL Ratio $\alpha$, Edge Dropping Ratio $P_{\text{ED}}$, and Node Dropping Ratio $P_{\text{ND}}$ on the ISCX-Tor Dataset (The shallow lines are attained by smoothing the dark plotted lines)}
	\label{Torsensitivity}
\end{figure*}

\begin{figure*}[t]
    \centering
	\begin{subfigure}{0.24\linewidth}
		\centering
		\includegraphics[width=1.0\linewidth]{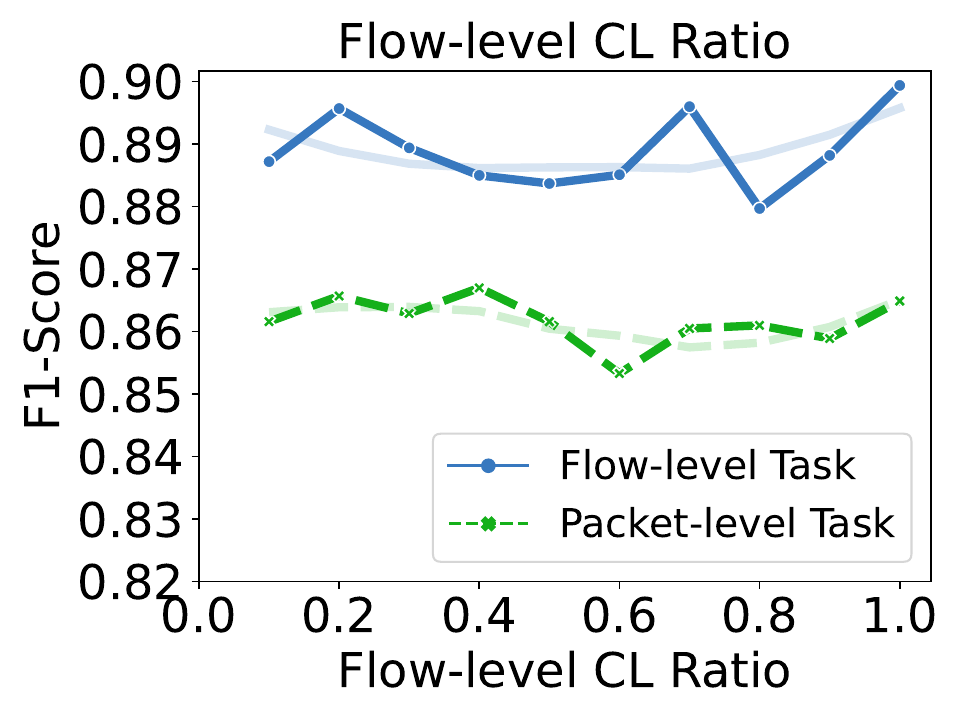}
		\caption{Flow-level CL Ratio $\beta$}
		\label{NonTorFCLR}
	\end{subfigure}
	\centering
	\begin{subfigure}{0.24\linewidth}
		\centering
		\includegraphics[width=1.0\linewidth]{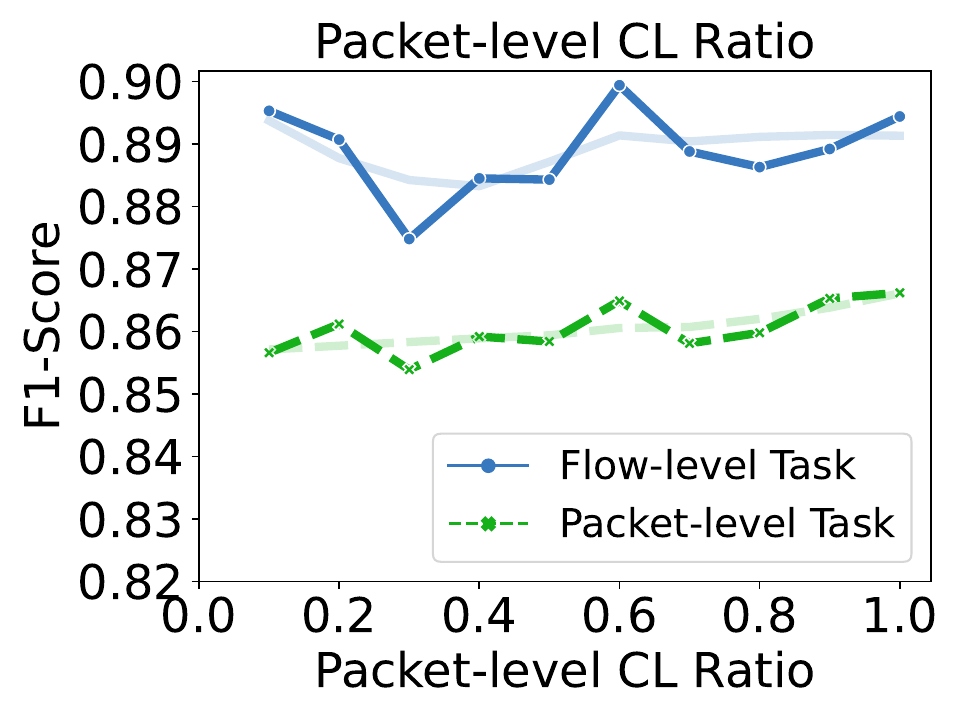}
		\caption{Packet-level CL Ratio $\alpha$}
		\label{NonTorPCLR}
	\end{subfigure}
	\centering
	\begin{subfigure}{0.24\linewidth}
		\centering
		\includegraphics[width=1.0\linewidth]{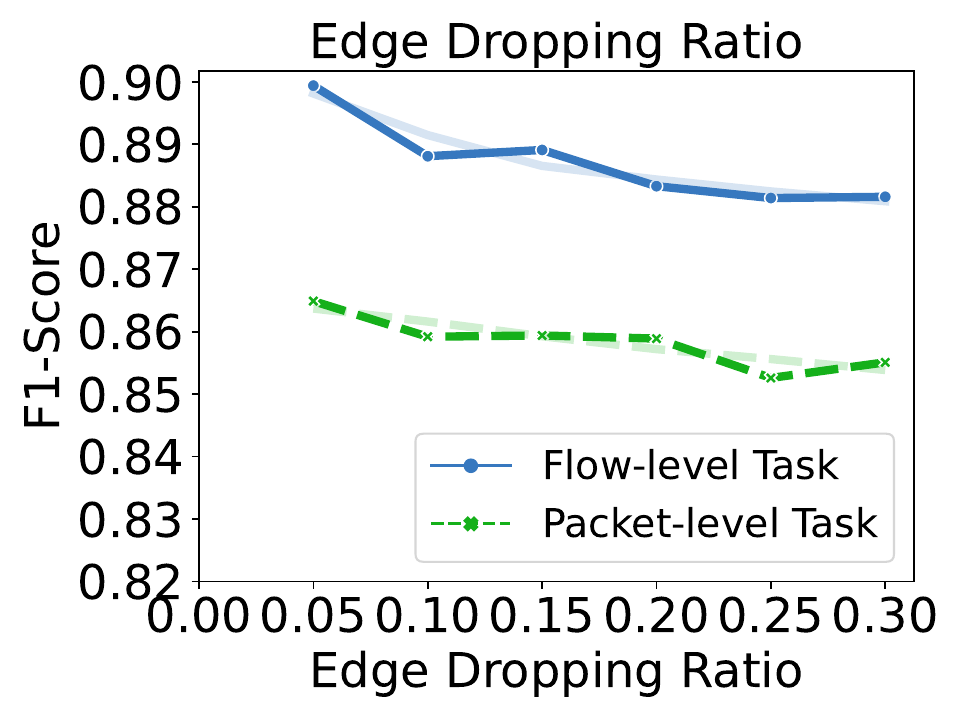}
		\caption{Edge Dropping Ratio $P_{\text{ED}}$}
		\label{NonTorEDR}
	\end{subfigure}
        \centering
	\begin{subfigure}{0.24\linewidth}
		\centering
		\includegraphics[width=1.0\linewidth]{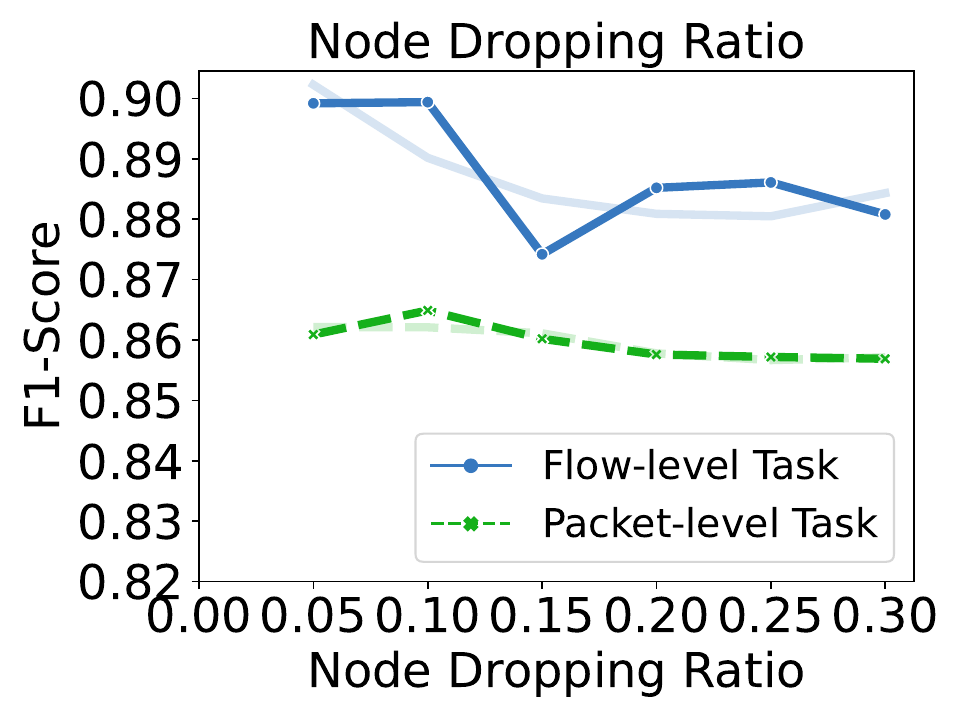}
		\caption{Node Dropping Ratio $P_{\text{ND}}$}
		\label{NonTorNDR}
	\end{subfigure}
	\caption{Model Sensitivity Analysis w.r.t. Flow-level CL Ratio $\beta$, Packet-level CL Ratio $\alpha$, Edge Dropping Ratio $P_{\text{ED}}$, and Node Dropping Ratio $P_{\text{ND}}$ on the ISCX-NonTor Dataset (The shallow lines are attained by smoothing the dark plotted lines)}
	\label{NonTorsensitivity}
\end{figure*}

In this section, we give more sensitivity analysis results on the ISCX-NonVPN, ISCX-Tor, and ISCX-NonTor datasets. The results are shown in Figure~\ref{NonVPNsensitivity}, \ref{Torsensitivity}, and \ref{NonTorsensitivity}, respectively.

\section{Hyper-parameters}
\label{sec:appendix_hyper}

We give hyper-parameters used in this paper in Table~\ref{tab:hyper-parameters}.

\begin{table*}[t]
  \caption{Hyper-parameters}
  \label{tab:hyper-parameters}
  \begin{tabular}{c|c|c|c|c}
    \toprule
    Datasets & ISCX-VPN & ISCX-NonVPN & ISCX-Tor & ISCX-NonTor \\
    \midrule
    Batch Size
    & 16 & 102 & 32 & 102 \\
    Gradient Accumulation
    & 1 & 5 & 1 & 5 \\
    Epoch
    & 20 & 120 & 100 & 120 \\
    Learning Rate (Max, Min)
    & (1e-2, 1e-4) & (1e-2, 1e-5) & (1e-2, 1e-4) & (1e-2, 1e-4) \\
    Label Smoothing
    & 0.0 & 0.01 & 0.0 & 0.0 \\
    Warm Up
    & 0.1 & 0.1 & 0.1 & 0.1 \\
    GNN Dropout Ratio
    & 0.0 & 0.1 & 0.0 & 0.2 \\
    LSTM Dropout Ratio
    & 0.0 & 0.15 & 0.0 & 0.1 \\
    Embedding Dimension
    & 64 & 64 & 64 & 64 \\
    Hidden Dimension
    & 128 & 128 & 128 & 128 \\
    PMI Window Size
    & 5 & 5 & 5 & 5 \\
    \midrule
    Edge Dropping Ratio $P_{\text{ED}}$
    & 0.05 & 0.05 & 0.05 & 0.05 \\
    Node Dropping Ratio $P_{\text{ND}}$
    & 0.1 & 0.1 & 0.1 & 0.1 \\
    Packet Dropping Ratio $P_{\text{PD}}$
    & 0.6 & 0.6 & 0.6 & 0.6 \\
    Temperature $\tau$
    & 0.07 & 0.07 & 0.07 & 0.07 \\
    Flow-level Contrastive Loss Ratio $\beta$
    & 0.5 & 0.8 & 1.0 & 1.0 \\
    Packet-level Contrastive Loss Ratio $\alpha$
    & 1.0 & 0.4 & 0.4 & 0.6 \\
    \bottomrule
  \end{tabular}
\end{table*}

\section{Dataset Statistics}
\label{sec:appendix_dataset}

We give specific category divisions for each dataset in the following: 
\begin{itemize}
\item \textbf{ISCX-VPN}: VoIP, Streaming, P2P, File, Email, Chat. 
\item \textbf{ISCX-NonVPN}: VoIP, Video, Streaming, File, Email, Chat. 
\item \textbf{ISCX-Tor}: VoIP, Video, P2P, Mail, File, Chat, Browsing, Audio. 
\item \textbf{ISCX-NonTor}: VoIP, Video, P2P, FTP, Email, Chat, Browsing, Audio. 
\end{itemize}

As for category types of the ISCX VPN-NonVPN dataset, some files can be labeled as either “Browser” or other types like “Streaming.” So, we abandoned “Browser” and used the remaining six types. Note that we didn’t delete data but used the alternative types instead.

\begin{table}[H]
  \caption{Dataset Statistics}
  \label{tab:dataset_sta}
  \begin{tabular}{c|c|c|c}
    \toprule
    Dataset & \#Flow & \#Packet & \#Category \\
    \midrule
    ISCX-VPN
    & 1674 & 19282 & 6 \\
    ISCX-NonVPN
    & 3928 & 33838 & 6 \\
    ISCX-Tor
    & 1697 & 22888 & 8 \\
    ISCX-NonTor
    & 7979 & 68024 & 8 \\
    \bottomrule
  \end{tabular}
\end{table}

\end{document}